\documentclass[11pt]{article}
\PassOptionsToPackage{table}{xcolor}

\usepackage[final]{acl}

\usepackage[T1]{fontenc}
\usepackage[utf8]{inputenc}

\usepackage{times}
\usepackage{latexsym}
\usepackage{inconsolata}
\usepackage{microtype}
\usepackage{hyperref}
\usepackage{url}
\usepackage{booktabs}
\usepackage{xcolor}
\usepackage{longtable}
\usepackage{array}
\usepackage{lineno}
\usepackage[section]{placeins}
\usepackage[most]{tcolorbox}
\usepackage{cleveref}
\usepackage{graphicx}
\usepackage{adjustbox}
\usepackage{float}
\usepackage{verbatim}
\usepackage{enumitem}
\usepackage{stfloats}
\usepackage{caption}
\usepackage{threeparttable}
\usepackage{amsmath}
\usepackage{amssymb}

\definecolor{darkblue}{rgb}{0, 0, 0.5}
\hypersetup{colorlinks=true, citecolor=darkblue, linkcolor=darkblue, urlcolor=darkblue}

\global\long\def\defeq{\stackrel{\mathrm{{\scriptscriptstyle def}}}{=}}

\crefname{appendix}{Appendix}{Appendices}
\Crefname{appendix}{Appendix}{Appendices}

\title{A Contextual Measure of Ambiguity in Prescriptive Text}

\author{
 \textbf{Sally Dong*\textsuperscript{1}},
 \textbf{Luke Thorburn*\textsuperscript{2,1}},
 \textbf{Andrew Konya\textsuperscript{1}},
 \textbf{Michiel Bakker\textsuperscript{1}}
\\
 \textsuperscript{1}Massachusetts Institute of Technology,\\
 \textsuperscript{2}AI \& Democracy Foundation \\
 * Equal contribution
\\
 \small{
   \textbf{Correspondence:} \texttt{sallyqd@mit.edu}
 }
}

\begin{document}

\maketitle

\begin{abstract}

    Ambiguity in prescriptive text such as laws, contracts, rules, policies, or recommendations can be both desirable (e.g., if it enables a law to adapt to changing context, such as new technologies) and undesirable (e.g., if it leads to arbitrary discretion or conflict over the law's interpretation). To help navigate these trade-offs, it would be valuable to have methods for quantifying ambiguity that are predictive of such outcomes, but existing measures capture how text is perceived in isolation rather than how it is interpreted in context. In this paper, we propose quantifying ambiguity as the degree of disagreement among a set of stakeholders (``raters'') on how alternative prescriptive texts (``policies'') apply in each of a set of relevant scenarios (``cases''). Using two policy domains, we present preliminary analysis suggesting that, even when using simulated raters, this approach (i) correlates with real human disagreement, and (ii) is robust to spurious text features. We then apply the measure to existing deliberation datasets to describe the association between policy ambiguity and support for those policies. A Python package for our framework and all data used in this paper are available \href{https://github.com/sally-dong/DELTA-ambiguity}{on GitHub}.
    
\end{abstract}

\section{Introduction}

    Ambiguity or vagueness is a consequential consideration when drafting prescriptive text such as laws, contracts, or policies.
    Too much ambiguity, and it becomes difficult for citizens to understand the rules they must adhere to, while officials may have greater discretion to enforce rules selectively. For example, in the United States there is a ``void-for-vagueness'' doctrine in constitutional law which holds that criminal statutes may be void if they lack sufficient specificity, motivated by a view that ``men of common intelligence cannot be required to guess at the meaning of [an] enactment''~\cite{crs_vagueness_overview,aga1960void}. The stakes can be national in scale: the 1995 Quebec sovereignty referendum failed by barely a percentage point on a ballot question whose wording left genuine doubt about what a `Yes' vote would have mandated, and Canada's subsequent Clarity Act now empowers the House of Commons to judge whether the question in a proposed referendum is sufficiently clear before it is held~\cite{clarity_act}.
    
    But ambiguity can also be useful. Some degree of interpretive flexibility can allow legal rules to adapt to unforeseen circumstances, including technological or social changes that could not have been fully specified in advance. Ambiguity can also preserve discretion for judges, agencies, or other decision-makers tasked with applying general principles to heterogeneous cases. In political settings, ambiguity can facilitate compromise by allowing actors who agree on broad goals to defer disagreement over implementation details.
    
    As large language models (LLMs) are increasingly used to write and interpret policies, and have further been proposed as tools to augment democratic processes~\citep{tessler2024ai,konya2025using}, questions about how policy texts are interpreted and how ambiguity should be operationalized become ever more salient. Despite this, ambiguity remains difficult to formalize and measure. There are early signs that the problem is both tractable and consequential: \citet{hall2026can} find that scoring prediction-market contracts against a hand-built clarity rubric is enough to predict which contracts will later end up in dispute. But such rubrics must be engineered for each domain, and so far, a general, principled operationalization of ambiguity does not exist.
    
    \begin{figure*}[t]
    	\centering
    	\setlength{\fboxsep}{15pt}
    	\setlength{\fboxrule}{0.5pt}
    	\fbox{\includegraphics[width=\dimexpr\textwidth-2\fboxsep-2\fboxrule\relax]{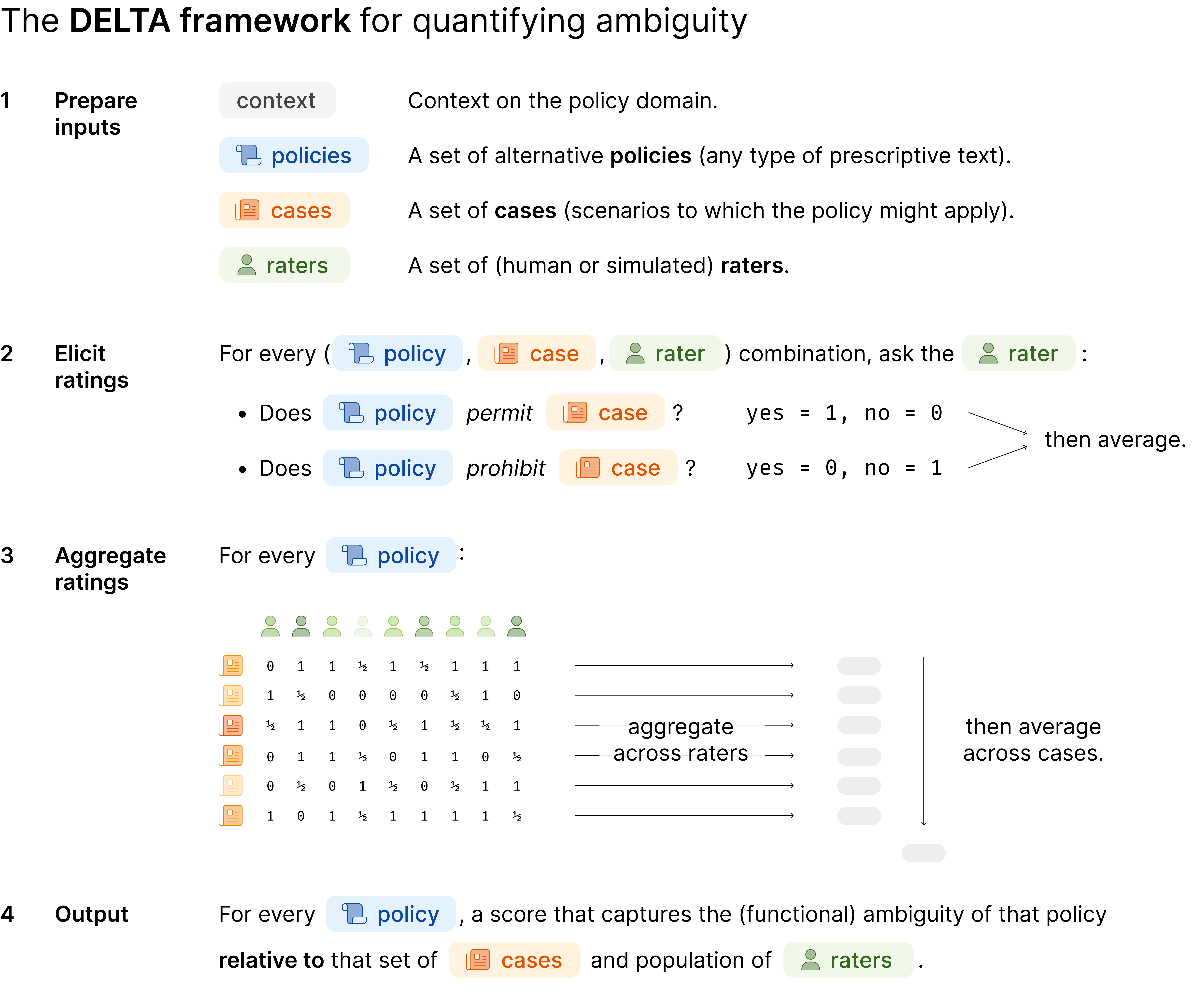}}
    	\caption{Visual intuition for the DELTA framework. Given context on a policy domain, a set of alternative policies, a set of \emph{cases} to which the policies might apply, and a set of \emph{raters}, each rater judges whether each policy permits and whether it prohibits each case, and the two answers are averaged into a single judgment. For each policy, disagreement is computed across raters for each case and then averaged across cases, yielding a score that captures the ambiguity of the policy relative to the chosen cases and population of raters. Cases and raters may be human-provided or, as in LLM-DELTA, generated and simulated with LLMs.}
    	\label{fig:delta-diagram}
    \end{figure*}

    In our paper, we introduce \emph{DELTA}, a framework for measuring ambiguity as interpretive divergence over consequences. DELTA evaluates a policy by pairing it with realistic application cases and asking a diverse set of raters whether the policy permits or prohibits each case. We incorporate LLMs to generate difficult evaluation cases, construct rater personas, and simulate judgments, producing a scalable approximation we call LLM-DELTA. %
    
    Across four datasets, we present preliminary validation of DELTA as a methodological framework and use it as a tool for substantive analysis.
    We first benchmark LLM-DELTA scores against DELTA scores obtained from a human-subject experiment, finding a substantial positive correlation in a single domain. In an adversarially constructed dataset, we show that DELTA recovers the intended distinction between consensus-inducing and disagreement-inducing policies, while direct LLM-as-a-judge ratings fail. We then apply the measure to two large-scale public discussions to test whether ambiguous statements are more likely to bridge disagreement. Contrary to the intuition that ambiguity facilitates compromise, we find null relationships: more ambiguous statements do not attract broader support, and they appear to soften strong agreement rather than convert opposition.

\section{Related work}

	Our approach to conceptualising and measuring ambiguity has connection to several large bodies of work from different disciplines. In this short survey, we aim to be reasonably comprehensive with respect to prior work on the narrow task of algorithmically quantifying degrees of ambiguity. We note also that use of terminology is not consistent in the literature; the general concept we refer to as ``ambiguity'' is sometimes also called ``vagueness'' \cite{aga1960void,lebanoff2018automatic}, ``specificity'' \cite{li2015fast} or ``concreteness'' \cite{yeomans2021concrete}, sometimes refers to different concepts (e.g., \cite{ellsberg1961risk}), and is sometimes decomposed in various ways, such as by distinguishing between \emph{subjective uncertainty} (where the interpreter finds a policy text unclear) and \emph{intersubjective disagreement} (where different interpreters disagree about how a policy applies to cases) \cite{farnsworth2010ambiguity,chen2023judgment}.

	Several approaches have been taken to quantifying ambiguity. One cluster of papers approaches the task as a binary classification, labelling texts as ambiguous/unambiguous or general/specific \cite{massey2014identifying,li2015fast,zhang2023using}, grounded in binary labels from human annotators. Slightly more granular approaches have humans score text on Likert scales capturing degrees of perceived ambiguity, and train models to predict those ratings \cite{yeomans2021concrete}. One notable paper introduced a dataset of human-labelled ``concreteness'' scores for about 40,000 English words and n-grams, rated on a 5-point Likert scale from abstract to concrete \cite{brysbaert2014concreteness}.

	A second line of work builds disagreement between interpreters into the conception of ambiguity itself, rather than treating ambiguity as a property of the text as perceived by an individual reader. In natural language inference (NLI), annotators judge whether a hypothesis follows from a premise, a task closely analogous to judging whether a policy permits a case. \Citet{pavlick2019inherent} showed that for many NLI examples, disagreement persists as more judgments are collected, and does not disappear when annotators are given additional context, indicating genuinely multiple plausible interpretations. \Citet{jiang2022investigating} taxonomized the causes of such disagreement into semantic uncertainty, annotator-specific assumptions, and artifacts of the task itself, distinctions that loosely parallel our decomposition of disagreement into intra-rater indeterminacy and inter-rater disagreement. \Citet{chen2023judgment} made a similar distinction operational in their \emph{Judgment Sieve} workflow, separately measuring \emph{ambiguity} (an individual's uncertainty about an item, often due to limited context) and \emph{disagreement} (divergence between individuals with different perspectives), and showing that targeting each with a matched intervention (adding context or deliberation, respectively) reduces both, whereas uniform interventions tend to trade one off against the other. This literature supplies the conceptual core of DELTA.

	Closest to our setting is work on rules written to govern AI systems. \Citet{he2025statutory} formalized the interpretive ambiguity of a rule as the entropy of compliance judgments across a panel of reasonable interpreters (instantiated as LLM judges), and adapted two mechanisms from the legal tradition to reduce it: iterative rule refinement at the rule-creation step, analogous to agency rulemaking, and prompt-based interpretive constraints at the rule-application step, analogous to canons of statutory construction. Approaching from the opposite direction, \citet{clifford2026open} studied the reconstruction of natural-language principles from preference data, and found that applying the resulting ``constitutions'' is itself ambiguous: different executors of the same constitution agree only 73\% of the time, leading them to conclude that principles can only be evaluated jointly with their interpreters.

	Most of the measures above are validated against judgments elicited for the purpose rather than against disagreements arising in practice. An exception is \citet{hall2026can}, who found that LLM-assigned scores on a clarity rubric predicted which prediction-market contracts were later disputed (out-of-sample AUC of roughly 0.75). But such rubrics must be engineered for each domain, and so far, a general, principled operationalization of ambiguity does not exist.

	The more automated versions of the algorithmic pipeline we introduce below include tasks of generating cases, generating personas (of raters), and for each policy-case pair, predicting whether a rater with a given persona would judge the case consistent with the policy. Versions of these tasks have been studied by others, for example work on case generation \cite{perez2022red,feng2023case,kuo2026botender}, persona generation \cite{park2022social,moon2024virtual,castricato2025persona}, case-based reasoning \cite{feng2023case,chen2025case}, and predicting the judgments of individual raters, both in annotation tasks with inherent disagreement \cite{zhang2021identifying,davani2022dealing,lee2023can} and in inferring opinions or votes on political and social issues \cite{argyle2023out,bakker2022fine,small2023opportunities,fish2024generative,gudino2024large}. We do not aim to advance state-of-the-art on these tasks with this paper, but expect that as performance on these tasks improves, those insights can be incorporated to improve our proposed ambiguity metric.

\section{Formalization} \label{sec:ambiguity}

    We begin with an abstraction: evaluating the consequences of a policy is equivalent to performing binary classification on its applicability to concrete scenarios. That is, given a \emph{case}, one must decide whether it is consistent with the policy (e.g., permitted by it, or compatible with the position it expresses). Ambiguity arises when this mapping from cases to binary decisions is ill-defined or contested. In the same spirit as \citet{farnsworth2010ambiguity}'s notion of intersubjective disagreement, we define the ambiguity of a policy as the expected disagreement among a diverse pool of raters regarding its consequences.

\subsection{The DELTA score}

    We operationalize the above view through the DELTA (\emph{Divergence Evaluation of Linguistic Text Ambiguity}) framework. Abstractly, given a policy, we construct hypothetical \emph{cases}, and poll \emph{raters} on whether they believe the case is consistent with, permitted by, or compatible with the policy. We then compute a \emph{disagreement} measure over these judgments, where a value of zero indicates perfect consensus and higher values indicate increasing division among raters. The \emph{DELTA score} of the policy is defined as an average of disagreement scores across the set of cases, with higher scores indicating greater ambiguity.

    DELTA is an intentionally flexible framework: users can tailor the ambiguity measurement through their choice of cases and stakeholder distributions. 
    Consequently, DELTA scores are relative and are meaningful only for comparing policies within a shared evaluation setup.
    Cases should be realistic yet engender debate regarding whether they are permitted or prohibited by the policy. Raters should be a diverse and broadly representative sample of domain stakeholders, so that their judgments capture a wide spectrum of viewpoints. This construction process reflects how policies are developed in practice: policymakers anticipate edge cases and potential conflicts that might arise during enforcement, so the policy language can be refined to preempt them.

    Formally, suppose we have $n$ policies $p_1, \dots, p_n$ in the same domain and wish to compare their ambiguity level. Assume we have already constructed cases $c_1, \dots, c_m$ and found raters $r_1, \dots, r_\ell$.
    Given policy $p_i$, case $c_j$, we ask each rater $r_k$ ``does $p_i$ permit $c_j$?'' and receive a binary answer $q_{ijk}$. Theoretically, if we then ask the rater ``does $p_i$ prohibit $c_j$?'', the answer $q'_{ijk}$ should be the complement, but in practice, this is not the case if $p_i$ is only tangentially related to $c_j$.
    As an example, consider the classroom AI policy ``students cannot use AI for their homework'' and case ``a student uses AI to quiz herself on course material''. The policy clearly does not prohibit the case, but it also does not affirmatively address it; raters may therefore feel unable to confidently say either that the policy permits the case or that it prohibits it.
    To capture such asymmetry, we average the answers to the positive framing and the complement of the negative framing to determine the \emph{revealed judgment}: $x_{ijk} \defeq (q_{ijk} + (1- q'_{ijk}))/2$. Each $x_{ijk}$ is interpreted as the probability that rater $r_k$ would issue a positive verdict.

    Next, let $\bar{x}_{ij}$ denote the mean of the revealed judgments across all raters. Observe that $2\,\bar{x}_{ij}\,(1-\bar{x}_{ij})$ is the probability that the verdicts of two raters drawn independently from the pool conflict (the Gini impurity of the pooled verdict), which is largest ($1/2$) when the pooled verdict is an even coin flip. We define the \emph{case disagreement score} of policy $p_i$ on case $c_j$ as this conflict probability normalized by its maximum:
    \[
    d_{ij} \defeq 4\,\bar{x}_{ij}\,(1-\bar{x}_{ij}),
    \]
    which is twice the Gini impurity, giving a range of [0,1], where the value is 0 under full consensus and 1 when the pooled verdict is an even coin flip.

    By the law of total variance, the score decomposes exactly into the two sources of interpretive conflict:
    \[
    d_{ij} = \underbrace{\frac{4}{\ell}\sum_{k=1}^{\ell} x_{ijk}(1-x_{ijk})}_{\text{intra-rater}} \;+\; \underbrace{4\,s_{ij}^2}_{\text{inter-rater}},
    \]
    where $s_{ij}^2$ is the population variance of the revealed judgments across raters. The first term captures \emph{intra-rater indeterminacy}: a rater with $x_{ijk}=0.5$ (e.g., because they cannot decide whether or how $p_i$ applies to $c_j$) conflicts even with an independent redraw of their own verdict. The second is \emph{inter-rater disagreement}: the additional conflict attributable to raters differing from one another. Both are symptoms of ambiguity, since a policy can fail to determine consequences either because no individual reader can tell what it implies, or because different readers reach incompatible conclusions, and the score credits both on a common probability scale.
    In particular, unanimous indeterminate judgments ($x_{ijk} = 0.5$ for every rater) yield the maximal $d_{ij} = 1$ through the intra-rater term alone.

    Finally, we average across all $m$ cases to arrive at the DELTA score for policy $p_i$:
    \[
    \textsc{DELTA}(p_i) \defeq \frac{1}{m} \sum_{j=1}^m 4\,\bar{x}_{ij}\,(1-\bar{x}_{ij}).
    \]
    The overall score takes value in $[0,1]$, with 0 indicating no and 1 indicating maximum ambiguity with respect to the chosen set of cases and raters.

\subsection{Approximation via language models}

    To make our framework tractable at scale, we develop a systematic method for LLMs to perform the two core tasks that would otherwise require inordinate human effort: (1) generating a diverse set of challenging synthetic evaluation scenarios for a given collection of policies, and (2) generating a diverse pool of synthetic raters and simulating their judgments.
    Full details of our prompts and model selections are presented in~\cref{appendix:model-and-prompt}.
    
\paragraph{Cases.}
    To generate challenging evaluation cases, we repeatedly sample pairs of policies and prompt a reasoning LLM (GPT-5) to produce diverse cases, using the pair as inspiration. The model is instructed to craft cases that comply with one policy in the pair while violating the other if possible. Since policies typically express similar goals, the LLM must exploit the nuanced differences between the pair to fulfill the request; as a result, the output highlights precisely the details that give rise to disagreements in interpretation, especially with respect to other texts besides the input pair. Asking for a batch of cases induces a minimum amount of contextual variety; on the other hand, to avoid redundancy in the underlying cause of disagreement, we sample per batch to keep in the final set.

\paragraph{Rater personas.}

    To construct a diverse pool of raters, we prompt a reasoning LLM (GPT-5) with the policy topic and request short descriptions of stakeholder identities in batches. To encourage viewpoint diversity, identities are generated in opposed pairs (e.g., on the policy topic of campus free-speech, a pair might be a university administrator versus a student activist).

    In preliminary testing, we found directly asking an LLM to adopt a rater identity and issue a binary judgment yields a narrow distribution of responses, especially in non-reasoning models. Prior work has shown that persona-conditioned LLMs can approximate some human response patterns but exhibit demographic misalignment and prompt sensitivity \cite{argyle2023out, aher2023using, santurkar2023whose}, and that richer agent state---e.g., natural-language reflection over memories and goals---can improve simulation fidelity \cite{park2023generative}. To elicit more differentiated perspectives, we introduce an intermediate step: the model first expands each identity into a brief paragraph describing that stakeholder’s values and concerns regarding the policy topic, serving as a lightweight reasoning trace that anchors subsequent judgments in a more coherent viewpoint. 
    
    The final prompt provided to the LLM therefore includes the rater identity, their generated reflections on the topic, and the judgment question in both positive and negative framing. We use a relatively lightweight LLM, Gemini-2.5-Flash at temperature 1, to carry out the judgments.

\subsection{Score convergence}
    \begin{figure}[h]
    	\centering
    	\includegraphics[width=\columnwidth]{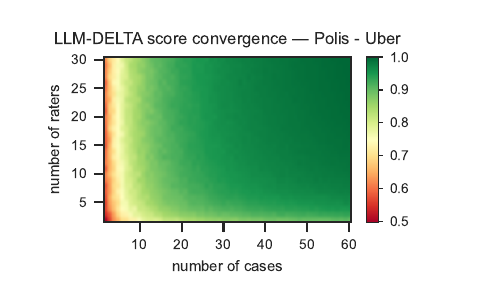}
    	\caption{Each cell in the heatmap reports the average Spearman correlation between LLM-DELTA scores computed from a random subset of $i$ cases and $j$ raters and reference scores computed using the full evaluation design.}
    	\label{fig:delta-convergence-uber}
    \end{figure}

    Repeated prompting of LLMs is known to yield diminishing returns in semantic diversity: newly generated cases or raters tend to resemble those already produced. Consequently, the marginal impact of additional samples on DELTA scores decreases over time, suggesting that a practical stopping criterion can be based on empirical convergence of the scores.

    We illustrate this using the Polis--Uber dataset introduced formally in~\Cref{subsec:datasets}. 
    For each policy in this dataset, we first compute a reference LLM-DELTA score using the full evaluation design, with 60 generated cases and 30 LLM raters. We then estimate how quickly these scores converge by recomputing LLM-DELTA scores on smaller random subsets of the full design:
    For each pair of subset sizes $(i,j)$, we repeat the subsampling procedure 100 times, each time sampling $i$ cases and $j$ raters, compute LLM-DELTA scores on this subset, and calculate the Spearman correlation between the resulting policy rankings and the full-sample reference rankings. The heatmap reports the average correlation across the 100 iterations.
    
    The results show rapid convergence (\cref{fig:delta-convergence-uber}). Correlations with the full-sample scores increase quickly as the number of sampled cases and raters grows, approaching high values well before reaching the full design of 60 cases and 30 raters. We note that this diagnostic establishes \emph{within-pool} rank stability---that subsampling from a fixed pool of generated cases and personas yields similar rankings to the full pool---rather than convergence to the broader distribution of cases and personas that further generations could produce. With this caveat, we use the diagnostic as a practical stopping criterion across datasets, typically with 30 raters and at most 50--60 cases as a balance between score stability and computational cost. In future applications, researchers can use the same diagnostic to select an appropriate number of cases and raters empirically.
    \Cref{appendix:convergence-human-experiment} reports the same diagnostic for our human subject study with similar results.

\section{Methods}

    In this section, we describe our hypotheses and approach to evaluating them.

\subsection{Hypotheses}

    We had two hypotheses about the validity of the LLM-DELTA measure of ambiguity. First, LLM-DELTA scores would better correlate with human-DELTA scores than would other baseline ambiguity measures (including LLM-as-a-judge, and a measure based on word-level concreteness scores). When comparing ambiguity measures on a set of policies, we focus on rank correlation because the DELTA score is a measure of relative rather than absolute ambiguity. Second, LLM-DELTA scores would be more robust than the baseline methods to ``adversarial'' examples: policies that appear unambiguous (because they are worded in a way that sounds decisive), but that are actually ambiguous (because they lead to significant disagreement among a given population of raters when applied to a given set of cases), or vice versa.

    We had one substantive hypothesis, tested by applying the measure,  that ambiguity in prescriptive text can help with consensus or agreement at the policy level by effectively postponing disagreement to future disputes about how the policy should be applied to individual cases. We hypothesize that this effect will be visible in real datasets as a positive relationship between ambiguity and various measures of consensus or \emph{bridging} agreement~\cite{ovadya2022bridging}.

\subsection{Datasets} \label{subsec:datasets}

    We worked with four disparate datasets, corresponding to multiple policy domains. The complete datasets can be found in \href{https://github.com/sally-dong/DELTA-ambiguity}{our project repository}.

\paragraph{Dataset 1: University speech.} 

    For benchmarking LLM-DELTA and other measures against a human-DELTA baseline, we used the domain of university policies related to freedom of expression, particularly pertaining to speech about harm or violence. We (ourselves) wrote 20 alternate policies, designed to span from the highly ambiguous (e.g., platitudes about maintaining a ``welcoming environment'') to highly unambiguous (e.g., modeled on the widely-used legal test for incitement). For cases, we paraphrased 20 real campus speech controversies from databases maintained by the Foundation for Individual Rights and Expression (FIRE) \cite{fire_deplatforming,fire_scholars_under_fire,fire_students_under_fire}. Both policies and cases can be found in \cref{app:human-study}.

\paragraph{Dataset 2: Adversarial.}

    For testing the robustness hypothesis, we developed a procedure for generating ``adversarial'' configurations of policies, cases, and rater personas (described in full in \cref{app:adversarial-procedure}). In brief, it begins with a minimal example of two policies, two raters, and four cases such that the raters agree on all cases under one policy but disagree on all cases under the other, due to where the cases sit on underlying axes of consensus and conflict. For the results reported below, we generated one adversarial setup modeled on the 1999 Australian republic referendum and scaled it up to 20 policies, 32 cases, and 30 rater personas.

\paragraph{Dataset 3: Pol.is.}

    Pol.is is an online platform for large-scale deliberation and opinion mapping conducted in partnership with governments or civic organizations interested in broad public participation.
    Given a policy topic, participants submit short position statements and vote on statements submitted by others. 

    Although the statements in Pol.is conversations are not formal policies, they are policy-relevant texts where participants express what they believe governments ought to do, and we therefore treat them as directional policy proposals. In the DELTA framework, cases can be understood as policy implementations or decisions that might be evaluated against the proposal. Under this interpretation, it is still meaningful to ask whether a statement is ambiguous; that is, whether different raters disagree about the concrete actions supported and ruled out by the statement.

    We use four publicly available Pol.is datasets~\cite{polis_open_data}. The first concerns how Taiwan should regulate app-based ride-hailing services, collected as part of the \emph{vTaiwan} participatory process that contributed to subsequent legislation~\cite{vtaiwan_info}. The remaining three are sub-topics drawn from Austria’s \emph{Klimarat} process, a national citizens’ assembly on climate policy, covering production and consumption, housing, and energy~\cite{paice2022polis_klimarat}. Full details are provided in \cref{appendix:polis}.

\paragraph{Dataset 4: Habermas experiment.}

    \citet{tessler2024ai} developed an AI system, the \emph{Habermas Machine}, to help small groups of citizens identify common ground on social and political issues. The system takes as input individual opinion statements on a select topic and generates \emph{collective statements} intended to reflect positions that the group could broadly endorse.
    
    The system was evaluated in a human experiment in the UK. Participants were assigned to groups of four or five and asked to deliberate on one of eight political topics (e.g., ``should the minimum wage be raised in the UK?''). To begin, participants reported their personal agreement with the topic statement on a 1--7 scale and wrote a private opinion statement. The Habermas Machine then used these individual statements to generate collective statements representing possible common ground within the group. Group members then rated each collective statement according to their own level of agreement on a 1--7 scale, and a refinement iteration followed.
    
    For each discussion topic, we aggregated Habermas-generated collective statements across groups for ambiguity evaluation. This yields eight topic-specific sets, each with approximately 70 entries; full details of the experiment and our dataset construction are provided in \cref{appendix:Habermas}.

\subsection{Baselines}

    For eliciting a human-DELTA ground truth, we conducted a human subject study where a politically-balanced sample of U.S. adults, recruited via Prolific, was asked to rate (policy, case) tuples from Dataset 1 described above. The consent form, instructions, and a sample of the rating interface are included in Appendix \ref{app:human-study}. Each participant rated all 20 cases against two policies and was paid \$6; the median completion time was 18 minutes. Each policy was shown to 30 raters in total, distributed evenly across self-identified Republicans, Independents, and Democrats, allowing us to estimate disagreement across a politically diverse group. This setup differs slightly from the LLM-DELTA setup used elsewhere in the paper where the same rater evaluates all policies against all cases, due to practical limitations of human-subject experiments. 

    As a lexical baseline, we use a dictionary-based measure from \citet{brysbaert2014concreteness}. The dictionary contains approximately 40,000 English words, each assigned a human-annotated concreteness score on a 1--5 scale. For each policy, we identify words that appear in the dictionary (consistently above 85\% of the full text) and compute the average concreteness score across those words. Since more concrete language is typically less ambiguous, we negate the resulting score so that higher values correspond to greater ambiguity.

    We further establish two direct LLM-as-a-judge measures of ambiguity~\citep{zheng2023judging}. In these baselines, an LLM is given a policy to be rated on a 1--7 scale. We use two prompt variants. The first prompt defines ambiguity in standard dictionary-style terms, with higher values indicating more perceived ambiguity, while the second is designed to more closely mirror the DELTA mechanism: the model is asked to imagine plausible scenarios and assess the degree to which people with diverse viewpoints might disagree about the applicability of policy in those settings.
    Full prompts and model choices are reported in \cref{appendix:model-and-prompt}.

\section{Results} \label{sec:analysis}

\subsection{Human-DELTA benchmark}

    \begin{figure}[h]
    	\raggedright
    	\includegraphics[width=\columnwidth]{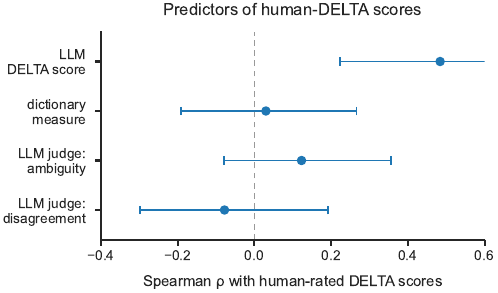}
    	\caption{Spearman correlations between four computational ambiguity measures and the human-DELTA scores for the first dataset of 20 university free-speech policies, with bootstrapped 95\% confidence intervals over 2000 iterations.}
    	\label{fig:human-DELTA-benchmark}
    \end{figure}

    \begin{figure}[h]
    	\raggedright
    	\includegraphics[width=\columnwidth]{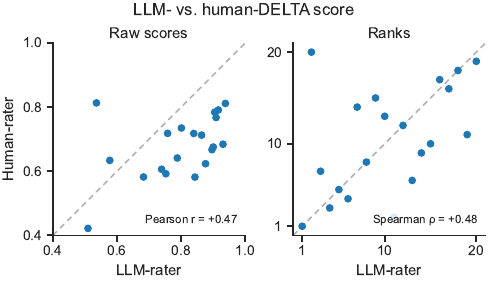}
    	\caption{Correlations between LLM-DELTA and human-DELTA on dataset 1, with raw scores shown on the left and rankings shown on the right. Each point represents one policy.}
    	\label{fig:LLM-human-DELTA-corr}
    \end{figure}
    
    \Cref{fig:human-DELTA-benchmark} reports Spearman correlations between human-DELTA scores and four alternative computational measures from the first dataset.
    Confidence intervals come from a stratified rater bootstrap: for each policy we resample raters with replacement separately within each party (Democrat, Independent, Republican) while preserving the original per-policy party distribution, and recompute the human-DELTA on each resample while holding the other variable fixed.
    
    Human-DELTA and LLM-DELTA scores are positively associated in accordance with our hypothesis (\cref{fig:LLM-human-DELTA-corr}), suggesting that the LLM-based procedure captures a similar form of interpretive disagreement to that observed in human judgments. This holds despite our not manually specifying the demographic or partisan characteristics of simulated raters, whereas the human experiment used raters balanced across Republicans, Independents, and Democrats.

    By contrast, the dictionary-based measure shows little relationship with human-DELTA, suggesting that DELTA does not simply capture surface-level word choice. The direct LLM-as-a-judge ambiguity rating fares somewhat better than the dictionary measure, but its association remains weak and far below LLM-DELTA's, likely because direct ratings are more sensitive to apparent vagueness or rhetorical style than to disagreement over downstream application.

    Finally, the LLM-as-a-judge disagreement measure performs poorly relative to LLM-DELTA, despite being prompted with a description of the DELTA mechanism. This indicates that merely asking a model to reason abstractly about possible disagreement is not sufficient to reproduce DELTA scores, as it lacks the scaffolding through concrete cases and diverse judgments.

\subsection{Adversarial robustness}

    \begin{figure}[h]
    	\centering
    	\includegraphics[width=\columnwidth]{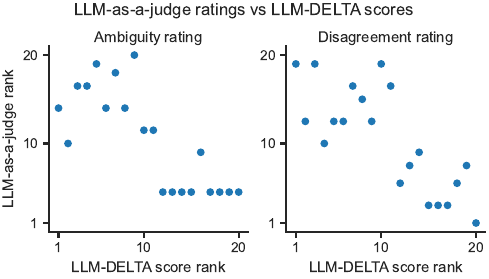}
    	\caption{Each point represents one policy in the adversarial dataset. Both LLM-as-a-judge ambiguity ratings and disagreement ratings rank the policies in nearly the reverse order of LLM-DELTA.}
    	\label{fig:adversarial}
    \end{figure}

    Recall that the adversarial policy set begins from a minimal example with two policy types: one that should elicit consensus among two raters, and is therefore unambiguous under DELTA, and one that should elicit disagreement, and is therefore ambiguous under DELTA. We then scale this construction to a full set of 20 policies, with ten policies of each type (and similarly scale up the two types of raters and four types of cases). DELTA successfully recovers this intended distinction in the richer evaluation design: it ranks the ten consensus-inducing policies as the ten least ambiguous policies.
    
    As shown in \cref{fig:adversarial}, both versions of the LLM-as-a-judge baselines rank the adversarial policies in nearly the reverse order of DELTA, revealing their lack of robustness. The adversarial setup was designed to decouple apparent ambiguity from realized disagreement along a latent axis; the reversal thus demonstrates that direct LLM-as-a-judge ratings fail to discover the pitfall, even when explicitly prompted to reason about disagreement.

\subsection{Relationship with bridging}

    Pol.is uses participant voting patterns to identify preference clusters. In the datasets we use, Pol.is identifies two primary opinion groups: participants within each group vote similarly on statements, while the two groups differ systematically. We use these algorithmically identified groups to examine the relationship between ambiguity and \emph{bridging}.

    Bridging is the ability of a statement to garner support from two opposing sides on a topic. Let $p_A$ and $p_B$ denote the agreement rates for a statement among participants in groups $A$ and $B$ respectively. We define the bridging score as $\min(p_A, p_B)$. This measure is conservative: a statement scores highly only if both groups agree with it at high rates. Statements that are strongly supported by one group but rejected by the other receive low bridging scores.
    Empirically, our definition closely tracks alternative \emph{cross-group alignment scores}, such as $p_A p_B$, used by Pol.is, and $p_A p_B + (1-p_A)(1-p_B)$, which is high when both groups tend to give the same response, whether agreement or disagreement. Our definition more clearly reflects shared endorsement: statements that can serve as common ground by attracting support from both opinion groups, rather than merely produce common rejection.

    \begin{figure}[h]
    	\centering
    	\includegraphics[width=\columnwidth]{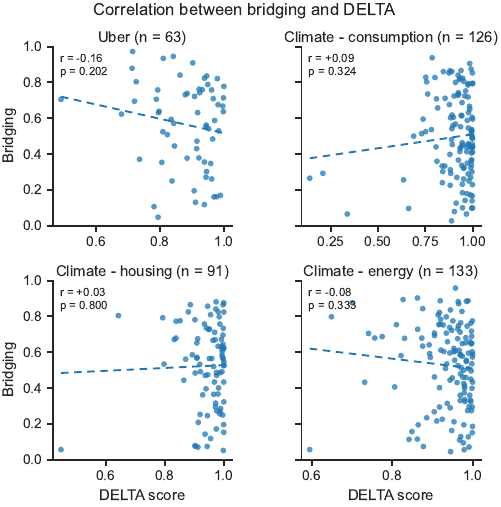}
    	\caption{LLM-DELTA and bridging show null or negative Pearson correlation across four Pol.is datasets, contrary to the intuition that ambiguity fosters broader support.}
    	\label{fig:bridging-delta}
    \end{figure}

    We find that ambiguity is not significantly associated with bridging in any of the four datasets at the 20\% engagement threshold shown in \cref{fig:bridging-delta}. Across the full range of thresholds examined in \cref{appendix:bridging-sweep}, no dataset shows a stable positive association---the direction the ambiguity-as-compromise hypothesis predicts. %

    One interpretation of this absence of a positive association is that in public deliberation, ambiguity does not necessarily create common ground by permitting interpretive latitude: For a statement to attract support from both opinion groups, it may need to articulate a position that is thoughtfully responsive to the underlying question. Ambiguous statements, by contrast, may fail to reassure either side about what the proposal would imply in practice and be seen as counterproductive. In this sense, bridging may depend less on allowing for multiple interpretations than on identifying concrete solutions that both groups can endorse.
    
    \begin{figure}[h]
    	\centering
    	\includegraphics[width=\columnwidth]{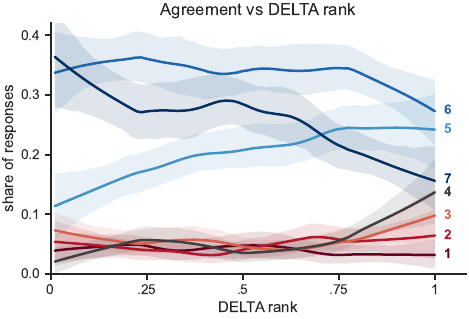}
    	\caption{Collective statements are ranked from least to most ambiguous according to DELTA scores. Lines show smoothed shares of participant agreement ratings on the 1--7 scale. 95\% confidence bands are obtained using 300 bootstrapping iterations.}
    	\label{fig:habermas-agreement}
    \end{figure}

    We complement this finding with an analysis of how ambiguity is associated with the degree of individual agreement using the Habermas dataset.
    Here, we consider only collective statements from groups with at least one participant initially supporting the discussion topic and at least one opposed, since those with no initial disagreement cannot meaningfully test opinion change.
    Within each topic, we rank statements by their DELTA score and normalize to $[0, 1]$, then examine how the distribution of participants’ agreement level varies as ambiguity increases. This within-question ranking allows us to compare patterns across topics without assuming that DELTA scores are similarly distributed.

    \Cref{fig:habermas-agreement} depicts the distributional shifts in participant agreement as a function of DELTA. Most notably, as DELTA increases, the proportion of ``strongly agree'' drops significantly, while ``slightly agree'' and neutral responses gain by a comparable combined amount; the entire disagree side changes only modestly. This pattern reinforces the Pol.is finding: Highly ambiguous statements do not appeal to the initially opposed to produce a bridging effect, but rather have the additional downside of softening the enthusiasm of the initially supportive.

\section{Conclusion}

We put forth a definition of ambiguity in prescriptive text as a divergence in interpretations of consequences. The DELTA framework operationalizes this view by making explicit the cases and stakeholder perspectives under which ambiguity is evaluated, enabling comparisons of policies beyond lexical properties alone. Our results further complicate a common view of ambiguity as politically useful: while ambiguity may sometimes enable elite bargaining or defer conflict, in the public deliberation settings we examined, it does not reliably produce common ground. Future work should improve rater simulation, refine case generation, benchmark the framework across additional domains and languages, and investigate whether DELTA can predict downstream implementation conflict in real policy settings.

\section{Limitations}

Our framework and findings come with several limitations.
First, LLM-DELTA depends on the capabilities and biases of the underlying language models used to generate cases, construct rater personas, and produce judgments. Although we observe a positive correlation between LLM-DELTA and human-DELTA on our university speech dataset (Spearman $\rho = 0.48$), this is a moderate rather than strong association, and it is established on a single domain with 20 policies. The extent to which LLM-simulated raters reproduce real interpretive disagreement in other policy domains, particularly those underrepresented in pretraining data, those involving non-English text, or those where stakeholders hold views that are systematically misrepresented in training corpora, remains an open empirical question. Prior work has documented that persona-conditioned LLMs exhibit demographic misalignment and prompt sensitivity, and our framework inherits these issues.

Second, DELTA scores are inherently relative: they are only meaningful for comparisons within a shared evaluation setup (pool of cases and raters). Scores cannot be directly compared across studies that use different case sets or rater distributions, and the choice of cases and personas embeds assumptions about which scenarios and viewpoints are worth representing. Our case generation procedure samples policy pairs to surface points of nuanced disagreement, but it may systematically miss forms of ambiguity that only become apparent across larger sets of policies or in interaction with cases that no policy author would have anticipated.

Third, our human benchmark is limited in scope. We recruited 30 raters per policy in a single domain (U.S. university speech), balanced across self-identified Democrats, Independents, and Republicans, but this captures only one axis of viewpoint diversity in one context. The human study also differs from the LLM setup in that each human rater evaluated only two policies rather than all 20, which may affect comparability. The confidence intervals in \cref{fig:human-DELTA-benchmark} hold LLM-DELTA scores fixed; so, for example, the relationship between LLM- and human-DELTA scores reflects variance in human judgments but not in LLM case generation, persona generation, or judgment sampling. We did not benchmark against human-DELTA in the Pol.is or Habermas datasets, so the validity of LLM-DELTA in those settings is supported only indirectly.

Fourth, the adversarial robustness result demonstrates that LLM-as-a-judge baselines fail on a setup we constructed precisely to expose their weaknesses. We do not claim that LLM-DELTA would prove similarly robust to adversarial constructions designed against it; such constructions are plausible and worth investigating in future work.

Finally, our bridging analysis is correlational and observational. We cannot identify whether the negative or null relationships between ambiguity and bridging reflect a causal property of ambiguous statements, selection effects in what kinds of statements get submitted to Pol.is, or features specific to online deliberation that may not generalize to formal policy drafting contexts. Our analysis also does not account for plausible confounds such as statement length, extremist rhetoric, or LLM-translation artifacts.

\paragraph{Risks.}
We see limited direct risks from this work. DELTA is a measurement tool rather than a generative one, and it does not produce policy text, rank stakeholders, or make recommendations about which policies should be adopted. The most plausible misuse would be treating DELTA scores as objective measures of policy quality or legitimacy when they are in fact relative measures dependent on the chosen evaluation design; we have tried to be explicit about this throughout. A secondary concern is that using LLM-simulated raters in place of human deliberation, if applied uncritically in real policymaking contexts, could lend a veneer of stakeholder consultation to processes that have not actually consulted stakeholders. We view LLM-DELTA as a tool to support human judgment in early-stage policy drafting and analysis, not a substitute for genuine democratic participation.

\section{Acknowledgments}

We thank Emma Cabal\'e, Nan Clement, Manuel W\"uthrich, and Ariel Procaccia for helpful discussions. We used AI in this project for coding, data analysis, and drafting and editing the paper.

\bibliography{custom}

\FloatBarrier
\appendix
\crefalias{section}{appendix}
\crefalias{subsection}{appendix}
\crefalias{subsubsection}{appendix}

\renewcommand{\thefigure}{\thesection.\arabic{figure}}
\renewcommand{\thetable}{\thesection.\arabic{table}}
\setcounter{figure}{0}
\setcounter{table}{0}

\section{Human subject study}

\label{app:human-study}

We wrote 20 policies on the topic of campus free speech, designed to span from the highly ambiguous to highly unambiguous. For cases, we paraphrased 20 real campus speech controversies from databases maintained by the Foundation for Individual Rights and Expression (FIRE) \cite{fire_deplatforming,fire_scholars_under_fire,fire_students_under_fire}. 

\subsection{Policies}

\begin{enumerate}
    
    \item The university prohibits acts of physical violence or force on campus. Expression alone, in the absence of physical conduct or a true threat, does not violate this policy.
    
    \item The university prohibits expression that constitutes a true threat: a serious expression of intent to commit unlawful violence against a specific person or group.
    
    \item The university prohibits expression that is both (a) directed at inciting imminent lawless action and (b) likely to produce such action.
    
    \item The university prohibits faculty and staff from advocating violence against identifiable persons while acting in their professional capacity. Expression outside that capacity is not covered.
    
    \item Instructors may not present, as established fact in their teaching, claims contradicted by the consensus of the relevant field. Contested interpretations and opinions remain permitted.
    
    \item The university prohibits expression that celebrates violence committed against a specific person. Expression about ideas, groups, or abstractions is not covered.
    
    \item Instructors may not express personal hostility toward identifiable individuals in required classes. The same expression in voluntary venues is not covered.
    
    \item The university prohibits expression that disparages an identifiable group on the basis of race, sex, religion, nationality, sexual orientation, or political affiliation.
    
    \item The university prohibits expression that a reasonable person would interpret as encouraging others to commit acts of violence.
    
    \item The university prohibits expression that creates a hostile environment for members of the campus community.
    
    \item The university is committed to maintaining a diverse, inclusive, and welcoming learning environment for all members of its community.
    
    \item The university permits criticism of public figures, including criticism delivered in harsh or intemperate terms.
    
    \item Members of the campus community are expected to engage with one another in a spirit of mutual respect and civility.
    
    \item The university affirms the principles of academic freedom and the free exchange of ideas as essential to its educational mission.
    
    \item The university prohibits expression that undermines the dignity of members of the campus community.
    
    \item The university supports the right of every student to learn in an environment free from intimidation, harassment, and discrimination.
    
    \item The university values robust debate and the open contest of ideas, including ideas that members of the community may find objectionable.
    
    \item The university values courage, including the courage to speak and the courage to listen.
    
    \item The university affirms its commitment to truth.
    
    \item Members of the campus community are expected to conduct themselves in a manner befitting the institution.

\end{enumerate}

\subsection{Cases}

\begin{enumerate}

    \item An English instructor at the university shared an article on Instagram about the assassination attempt on Donald Trump, adding the comment: ``If you're gonna shoot, man, don't miss.'' The post was re-shared by the ``Libs of TikTok'' account on X (formerly Twitter), and gained widespread attention.
    
    \item A group of students, posting under a social media account called ``CanesCritters,'' uploaded a video for a Digital Marketing class assignment in which a Black student is compared to a gorilla.
    
    \item A professor in the Department of Asian and Middle Eastern Studies at the university, criticized the idea of free speech on campus, associating it with the rise of fascism, and mocked those who promote debate over resistance, saying the moral high ground is ``for those who wish to drown in blood.''
    
    \item A student, in a private group chat, described the January 6 Capitol breach as ``a peaceful occupation of the capitol and our people were provoked with tear gas and flash bangs!''
    
    \item A part-time math instructor modified the state's mandatory Oath of Allegiance to align with her pacifist beliefs. The standard oath requires employees to ``support and defend'' the U.S. and California Constitutions ``against all enemies, foreign and domestic.'' She inserted the word ``nonviolently'' before ``support and defend,'' crossed out ``swear,'' and circled ``affirm.'' Previously, similar modifications had been accepted without issue during her employment with public schools.
    
    \item An electrical engineering and computer science professor, author of a 1976 book called `The Hoax of the Twentieth Century' which denies the systematic extermination of six million Jews during World War II, publicly endorsed Iranian President Mahmoud Ahmadinejad's Holocaust denial statements, saying: ``I congratulate him\ldots{} It was about time an Islamic leader stated this publicly.''
    
    \item An emeritus professor of medicine publicly supported physician-assisted suicide and the removal of Terri Schiavo's feeding tube. (As context, Terri Schiava was a Florida woman who suffered severe brain damage in 1990 and was diagnosed as being in a persistent vegetative state. Her husband sought to have her feeding tube removed, arguing she would not have wanted to be kept alive in that condition; her parents fought to keep it in. The dispute went through years of court battles and became a national flashpoint in 2005, involving the Florida legislature, Governor Jeb Bush, Congress, and President George W. Bush, who signed emergency legislation aimed at keeping her alive. The feeding tube was ultimately removed and she died in March 2005.)
    
    \item A student government denied Turning Point USA's application for campus recognition, citing the group's ``misalignment'' with university values, ``confrontational language throughout the TPUSA website,'' the national organization's ``Professor Watchlist,'' and the group's ``actions of encouraging a growth in conflict rather than fostering communication.''
    
    \item A psychology professor posted on Facebook in response to Charlie Kirk's assassination: ``Me thinks the word `karma' is appropriate. Sad all the way around.'' In a second post, she wrote: ``But when we tell y'all that statistically \ldots{} White American men are the most dangerous animals on the planet we're wrong. Let's not be hasty they say \ldots{}'' She then deleted the posts.
    
    \item A psychology professor, when asked during a social psychology class about her views on NFL quarterback Colin Kaepernick's national anthem protests against police violence, said she ``would kill him.''
    
    \item A student produced a ``visual thesis'' that focuses on Islamic extremism and criticized violence and bigotry against Israel.
    
    \item A philosophy professor posted on Facebook: ``I want the entire world to burn until the last cop is strangled with the intestines of the last capitalist, who is strangled in turn with the intestines of the last politician.''
    
    \item Nine students, including two student journalists, participated in an unapproved ``die-in'' protest on campus to object to what they saw as the university administration's failure to acknowledge faculty and staff safety concerns regarding its COVID-19 reopening plan.
    
    \item A video circulated on Snapchat and Twitter showing an incoming student and two others laughing as one of them knelt on the chest of another who was saying ``I can't breathe.''
    
    \item A math instructor wrote in a Facebook exchange that there were ``no safe spaces in my classes'' for students in the country illegally, and said he would report such students to Immigration and Customs Enforcement (ICE).
    
    \item A professor of industrial and systems engineering, public policy, and economics displayed a ``Thin Blue Line'' flag, commonly associated with the ``Blue Lives Matter'' movement, on his office door (``blue lives'' refers to the lives of police officers). He said he aimed to present an alternative viewpoint and support law enforcement, asserting that ``blue lives protect black lives.''
    
    \item A professor of politics and global studies posted tweets about a Las Vegas mass shooting suggesting that ``white male entitlement'' was a contributing factor to mass shootings.
    
    \item A part-time anthropology lecturer was involved in a confrontation with members of the campus College Republicans during a counter-protest to a pro-Palestinian event. The College Republicans accused the lecturer of attempting to grab a student's sign and striking another student who intervened. The lecturer denied these allegations, asserting his commitment to nonviolence.
    
    \item The School of Public Affairs invited Condoleezza Rice to speak on civil rights. Rice supported the war in Iraq, the use of torture on detainees, and helped implement other Bush administration policies.
    
    \item Salman Rushdie, author of the 1988 novel The Satanic Verses, was invited by a college to deliver a commencement address. Rushdie had recently made comments equating pro-Palestinian campus protests to support for a ``fascist terrorist group.''

\end{enumerate}

\subsection{Study}

The study was determined to be exempt from ethics review by the university review board.
We recruited participants using Prolific and served the survey on Qualtrics.
Each participant rated all 20 cases against two policies and was paid \$6; the median completion time was 18 minutes. Each policy was shown to 30 raters in total, distributed evenly across self-identified Republicans, Independents, and Democrats, allowing us to estimate disagreement across a politically diverse group.

We used two attention check questions in the survey. They are presented as another case to rate per policy, and the order of cases is randomized. Seven percent of respondents failed at least one attention check; subsequently, we excluded their responses and recruited replacements.

\begin{figure*}[t]
  \centering
  \setlength{\fboxsep}{0pt}%
  \setlength{\fboxrule}{0.5pt}%
  \fbox{\includegraphics[width=\textwidth]{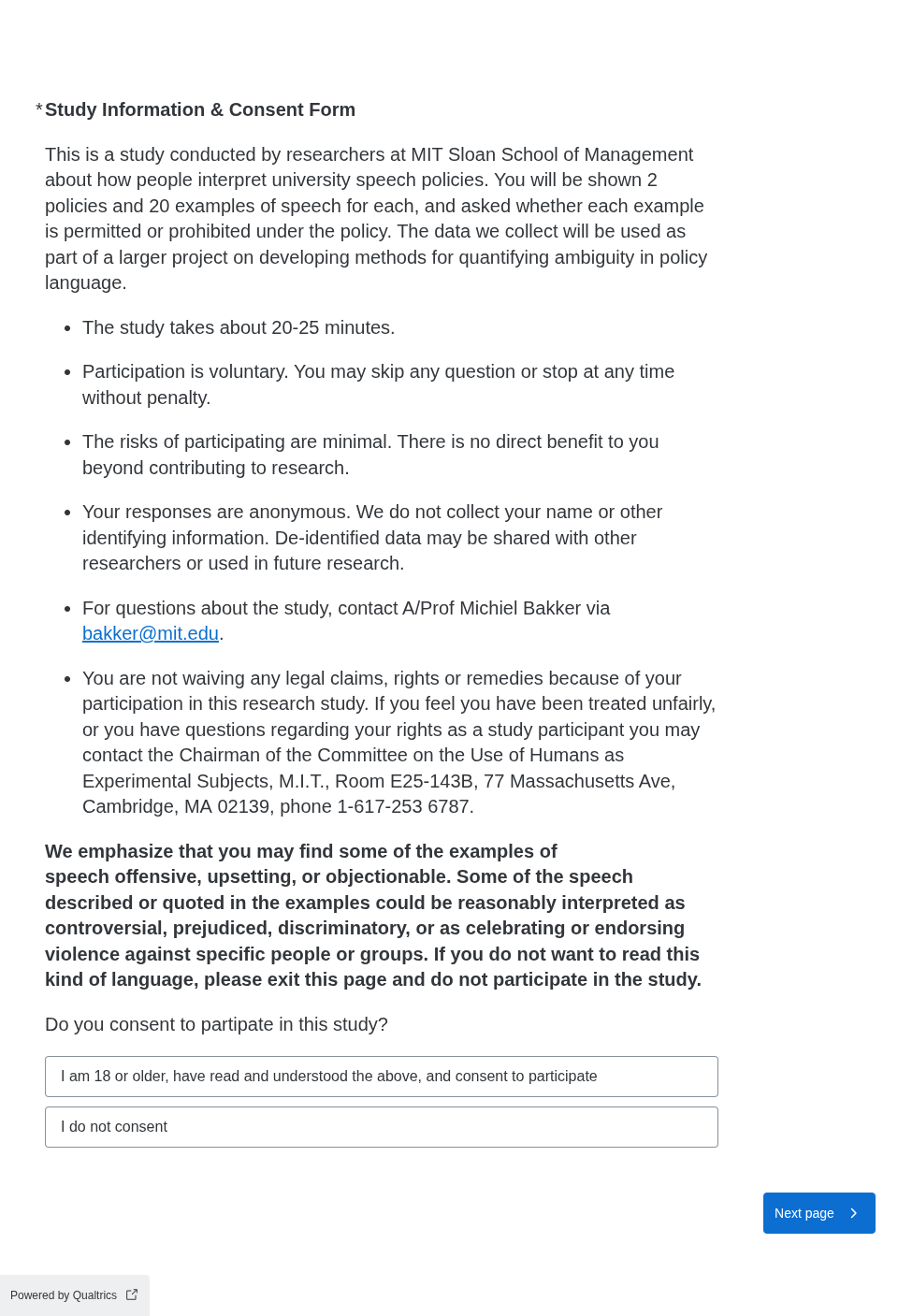}}
  \caption{The consent form used in the human subject study. Note that payment details had already been disclosed via the Prolific interface.}
  \label{fig:consent}
\end{figure*}

\begin{figure*}[t]
  \centering
  \setlength{\fboxsep}{0pt}%
  \setlength{\fboxrule}{0.5pt}%
  \fbox{\includegraphics[width=0.9\textwidth]{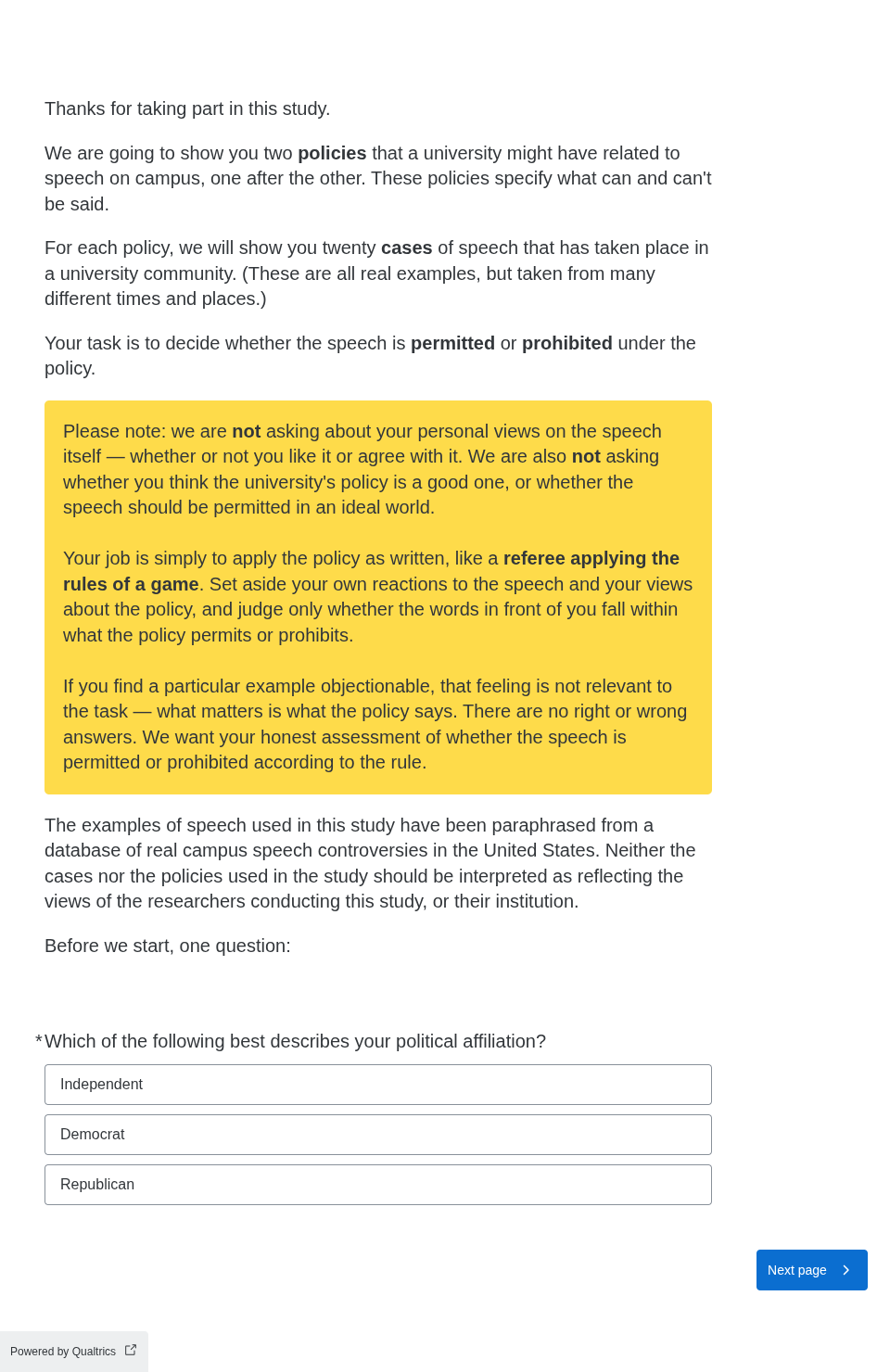}}
  \caption{Instructions given to participants in the human subject study at the beginning.}
  \label{fig:instructions}
\end{figure*}

\begin{figure*}[t]
  \centering
  \setlength{\fboxsep}{0pt}%
  \setlength{\fboxrule}{0.5pt}%
  \fbox{\includegraphics[width=\textwidth]{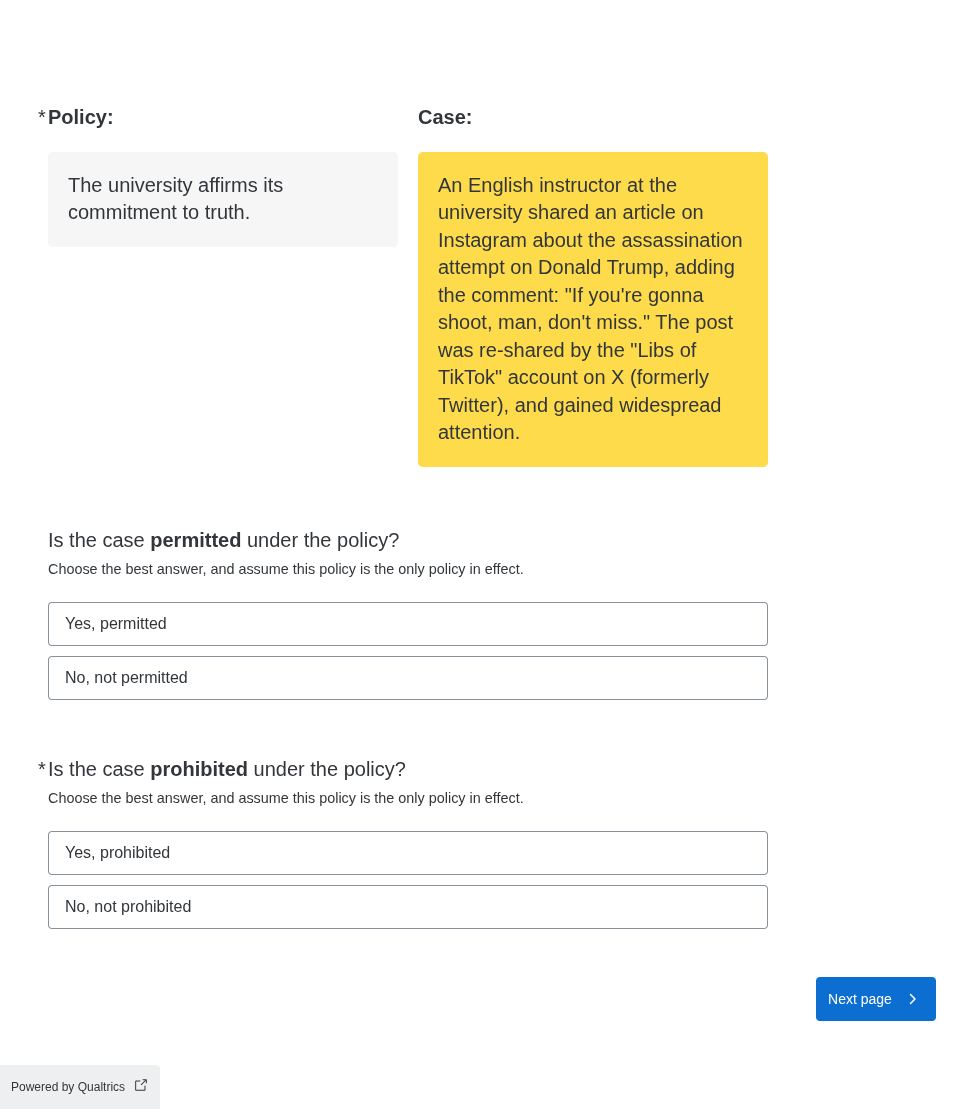}}
  \caption{An example of the rating task assigned to participants in the human subject study.}
  \label{fig:task}
\end{figure*}

\begin{figure*}[t]
  \centering
  \setlength{\fboxsep}{0pt}
  \setlength{\fboxrule}{0.5pt}
  \fbox{\includegraphics[width=\textwidth]{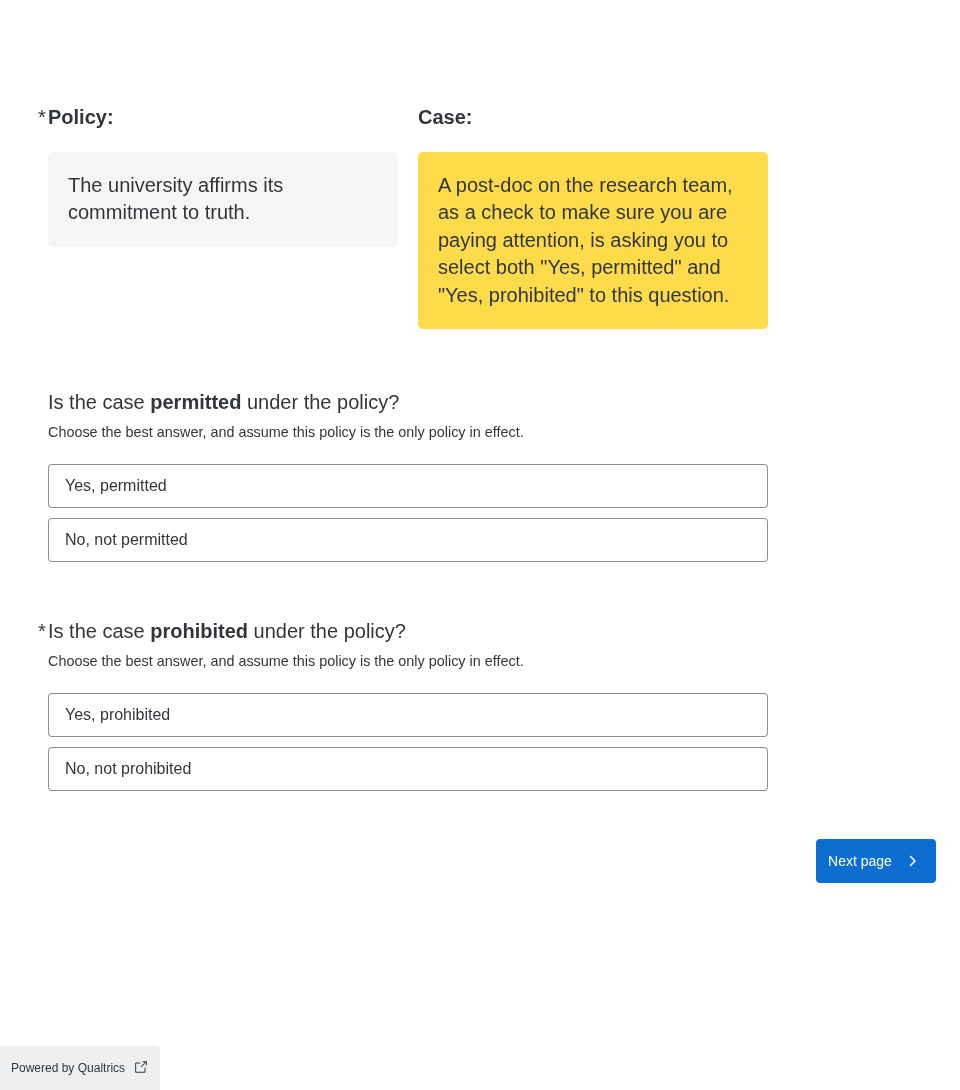}}
  \caption{An example of an attention check during the human subject study. Each participant was shown two such checks in total, and failing any led to the exclusion of their entire response from analysis.}
  \label{fig:attention-check}
\end{figure*}

\subsection{Score convergence}
\label{appendix:convergence-human-experiment}

\Cref{fig:delta-convergence-uber} in the main text reports the case-and-rater convergence diagnostic for the Polis--Uber dataset, whose cases are LLM-generated. \Cref{fig:delta-convergence-human-experiment} repeats the diagnostic on the human subject study dataset with hand-crafted policies and cases. 
We compute reference LLM-DELTA scores for the 20 policies using the full design of 20 cases and 30 LLM raters, and each cell reports the mean Spearman correlation between the reference ranking and rankings recomputed on 100 random subsamples of $i$ cases and $j$ raters.

The pattern matches the main text. Rankings stabilize well before the full design is reached: the mean correlation is $0.87$ with half the cases and a third of the raters (10 of each), and exceeds $0.9$ from roughly 8 raters and 15 cases.

\begin{center}
	\includegraphics[width=\columnwidth]{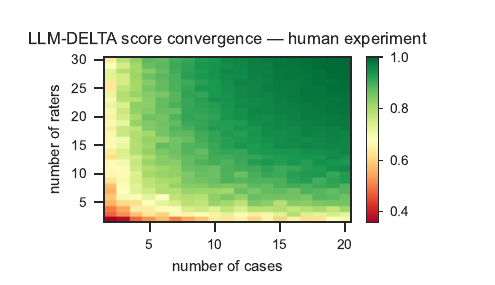}
	\captionof{figure}{Each cell reports the average Spearman correlation between LLM-DELTA scores computed from a random subset of $i$ cases and $j$ raters and reference scores computed using the full design of 20 manually curated cases and 30 LLM raters, on the campus speech dataset (100 subsamples per cell).}
	\label{fig:delta-convergence-human-experiment}
\end{center}

\begingroup\renewcommand{\FloatBarrier}{}
\section{Procedure for generating adversarial examples}
\label{app:adversarial-procedure}

    In this section, we describe the procedure used to generate the adversarial dataset referenced in \cref{subsec:datasets}. The goal of the procedure is to produce small, structured configurations of policies, cases, and rater personas in which two conceptions of ambiguity will lead to different assessments of the policies: `surface ambiguity', the impression that a policy is vague, underspecified, or open-textured when read in isolation (as measured by, e.g.,  LLM-as-a-judge), and `operational ambiguity': the disagreement that arises between interpretive communities when the policy is applied to concrete cases (measured, e.g., by LLM-DELTA).

\subsection{Overview}

    Each adversarial example consists of two policies, four cases, and two rater personas representing two interpretive groups. The two policies are designed to invert the relationship between surface and operational ambiguity:

    \begin{itemize}
        \item \textbf{Policy A} (the \emph{consensus} policy) is worded so as to \emph{sound} ambiguous---it uses open-textured, evaluative, or rhetorical language---but yields agreement between the two groups on every sampled case.
        \item \textbf{Policy B} (the \emph{conflict} policy) is worded so as to \emph{sound} concrete---it uses explicit, operational, or administrative language---but yields disagreement between the two groups, because its central terms encode a contested political, moral, institutional, or interpretive cleavage.
    \end{itemize}

    A surface-level ambiguity judge---whether an LLM-as-a-judge prompted to rate vagueness on a Likert scale, or a lexical measure based on concreteness---is expected to rate Policy A as more ambiguous than Policy B. An operational measure such as LLM-DELTA, which estimates ambiguity from disagreement among diverse raters about how the policy applies to cases, should rank the two policies in the opposite order. The adversarial dataset therefore functions as a discriminating test between the two families of measures.

\subsection{Latent dimensions}

    The construction relies on identifying two latent dimensions within a chosen policy domain.

    \paragraph{Consensus axis $C(\cdot)$.} A feature of cases that both groups classify in the same way, even though they may disagree on adjacent normative questions. Policy A is written to key onto this axis: although its language sounds vague, both groups read it as picking out the same feature of the cases.

    \paragraph{Conflict axis $H(\cdot)$.} A feature of cases whose meaning is contested between the groups---typically a term that is operationally clear on its face but interpreted differently across interpretive frameworks. Policy B is written to key onto this axis: although its language sounds concrete, the two groups disagree about whether particular cases satisfy it.

    The four cases are then drawn as a cross-product of the two axes. Each case takes one of two values along $C$ and one of two values along $H$, yielding the target judgment matrix shown in \cref{tab:adversarial-matrix}. Policy A produces uniform Y/Y or N/N judgments across the two groups (driven by $C$); Policy B produces split judgments (driven by $H$).

    \begin{table*}[t]
        \centering
        \small
        \begin{tabular}{lcccc}
            \toprule
                & \multicolumn{2}{c}{\textbf{Policy A}} & \multicolumn{2}{c}{\textbf{Policy B}} \\
                & \multicolumn{2}{c}{\emph{surface-ambig, operationally unambig}} & \multicolumn{2}{c}{\emph{surface-unambig, operationally ambig}} \\
                \cmidrule(lr){2-3} \cmidrule(lr){4-5}
            Case & Group 1 & Group 2 & Group 1 & Group 2 \\
            \midrule
            Case A ($C{=}Y$, $H$ favours G1) & Y & Y & Y & N \\
            Case B ($C{=}N$, $H$ favours G1) & N & N & Y & N \\
            Case C ($C{=}Y$, $H$ favours G2) & Y & Y & N & Y \\
            Case D ($C{=}N$, $H$ favours G2) & N & N & N & Y \\
            \bottomrule
        \end{tabular}
        \caption{Target judgment matrix. Each case takes a value on the consensus axis $C$ (shared by both groups) and on the conflict axis $H$ (interpreted differently by the two groups); the notation ``$H$ favours G1'' (resp.\ ``G2'') denotes the value of $H$ on which Group 1's (resp.\ Group 2's) reading of Policy B leads to Y. Y denotes that a group judges the case to be permitted by, or consistent with, the policy; N denotes the opposite.}
        \label{tab:adversarial-matrix}
    \end{table*}

\subsection{Construction procedure}
\label{subapp:adversarial-procedure-steps}

    To generate an example in a target domain, we follow nine steps.

    \begin{enumerate}
        \item \textbf{Choose a domain} with both (i) recognisable implementation distinctions that different stakeholders agree on, and (ii) contested concepts that different stakeholders interpret differently.
        \item \textbf{Define two groups} that plausibly partition the relevant population along an interpretive cleavage rather than along an outcome preference. Each group is specified by what it values, how it interprets key terms in the domain, and what kinds of implementations it sees as legitimate or illegitimate. Group labels should reflect interpretive stance rather than partisanship or desired outcome.
        \item \textbf{Identify a consensus axis} $C$ on which the two groups predictably agree. The axis should be a relatively concrete feature of cases even though the policy describing it will be worded vaguely.
        \item \textbf{Identify a conflict axis} $H$ on which the two groups predictably disagree, corresponding to a real political, moral, institutional, or interpretive cleavage in the domain. Each group should have a principled reading of the contested term.
        \item \textbf{Write Policy A} so that it tracks $C$ but uses surface-ambiguity cues such as open-textured evaluative adjectives, rhetorical contrasts, and value-laden but underspecified language.
        \item \textbf{Write Policy B} so that it tracks $H$ but uses surface-concreteness cues such as enumerated criteria, named harms, explicit institutional objectives, rights, or measurable-sounding outcomes.
        \item \textbf{Generate the four cases} as the cross-product of $C$ and $H$. Cases should be ecologically realistic---drawn from the kinds of proposals, programs, or incidents that actually arise in the domain---rather than artificial edge cases or technical loopholes.
        \item \textbf{Explain each judgment.} For every cell in the matrix, give a one- or two-sentence rationale for why each group reaches its Y or N verdict. The exercise serves as a verification step: if a cell's rationale is forced or implausible, the policy wording, group definition, or case is revised.
        \item \textbf{Verify the matrix} against \cref{tab:adversarial-matrix}. Examples that intentionally depart from the standard matrix are flagged explicitly and the departure is justified.
    \end{enumerate}

\subsection{Rater personas}

    Each group is operationalised as one or more rater personas: third-person descriptions of how a person within that interpretive community thinks about the domain, what they value, and how they interpret the contested terms in Policy B. Personas are written without self-reference to their group and use plausible role labels rather than outcome labels. At the minimal scale, one persona per group is produced. When more personas are added (see \cref{subapp:adversarial-scaling}), the role, vocabulary, and specific concerns are varied while the underlying interpretive stance is preserved, so that within-group stylistic variation can be distinguished from between-group disagreement.

    When real respondents are used in place of LLM personas, groups are derived empirically from a small set of anchor items diagnostic of the interpretive commitments, using rule-based scoring, latent-class analysis, or clustering. Groups are validated by checking that within-group agreement on anchor items is high, between-group disagreement is high, and assignments are stable on held-out items.

\subsection{Quality criteria}

    A candidate example is accepted only if it satisfies the following criteria.

    \begin{itemize}
        \item \emph{Surface plausibility.} Policy A reads as vague or open-textured to an unprimed reader, and Policy B reads as concrete or operational.
        \item \emph{Operational behaviour.} On the sampled cases, Policy A produces inter-group agreement and Policy B produces inter-group disagreement.
        \item \emph{Charitable groups.} Both groups have principled, internally consistent reasons for their judgments. Neither is portrayed as irrational, malicious, or caricatured.
        \item \emph{Ecological realism.} The cases are recognisable as plausible members of the domain---close to real policy instruments, programs, or disputes --- and not dependent on obscure technical loopholes, semantic tricks, or unrealistic hypotheticals.
        \item \emph{Boundary distance.} The Y/Y and N/N cases under Policy A are far enough from the consensus boundary that accidental within-policy-A disagreement is unlikely.
    \end{itemize}

\subsection{Scaling within a domain}
\label{subapp:adversarial-scaling}

    The default output of the procedure is a minimal set of two policies, four cases, and two rater personas. Three orthogonal directions of scaling are available within a single domain:

    \begin{enumerate}
        \item \emph{Paraphrase variation.} Multiple wordings of Policy A and Policy B are generated, preserving the consensus or conflict axis but varying surface vocabulary. Paraphrases test whether the operational pattern is robust to wording. They are not treated as independent conceptual examples.
        \item \emph{Cell variation.} Additional realistic cases are generated within each cell of the matrix. This produces a larger case set while preserving the four-cell structure. Cell counts need not be uniform; we favour ecologically realistic frequencies over balanced quotas.
        \item \emph{Rater paraphrase.} Additional personas are generated within each group, varying role and vocabulary while preserving the interpretive stance. This allows within-group stylistic variation to be distinguished from between-group disagreement.
    \end{enumerate}

\endgroup

\section{Pol.is dataset details} \label{appendix:polis}

We use four datasets from the Computational Democracy Project's open Pol.is data repository. Pol.is is a platform for large-scale online deliberation in which participants submit short statements and vote on statements submitted by others. The repository describes these datasets as anonymized human expressions in political contexts, gathered through facilitated conversations using the open-source Pol.is software. The repository also provides statement-level metadata, vote data, participant--vote matrices, conversation summaries, and aggregate statistics for each conversation. The data are distributed under a Creative Commons Attribution 4.0 license, with attribution to The Computational Democracy Project. Our research use is consistent with their intention.

Our analysis uses the statement texts from the provided \texttt{comments.csv} files. Each row corresponds to a submitted statement and includes a timestamp, comment identifier, author identifier, numbers of agrees and disagrees, a moderation field, and the statement text. The bridging analysis additionally uses the per-group vote tables included in the same data releases, which report, for each statement, the aggregate agree, disagree, and pass counts within each of the two opinion groups that Pol.is identifies by clustering participants' voting patterns. We use these published aggregates directly, rather than recomputing group assignments from the participant--vote matrices. We treat these statements as policy-relevant directional proposals: while they are not formal policy texts, they express claims about what governments or collective decision-makers should do. In the DELTA setup, cases are therefore interpreted as possible policy implementations or downstream decisions that could be evaluated against the proposal.

We use one conversation from the vTaiwan process and three conversations from Austria's Klimarat process. The vTaiwan conversation concerns the regulation of UberX and app-based ride-hailing services in Taiwan, held in 2015. 
The Computational Democracy Project describes this conversation as an engagement with the government of Taiwan as part of the vTaiwan participatory process that contributed to the regulation of Uber in Taiwan. The conversation summary describes the prompt as asking whether UberX, as an innovative ride-hailing service, should be required to follow the same laws as Taiwan's taxi or rental-car industries. In our analysis, we use the submitted statements from this conversation as the Taiwan ride-hailing subset.

The remaining three conversations are drawn from the Austrian Klimarat, a national citizens' assembly on climate policy, held in 2022. The Klimarat began after a petition concerning Austria's lack of a climate protection law received roughly 380,000 signatures, leading Parliament to fund a national citizens' assembly. The assembly selected 100 Austrians through stratified lottery, of whom 84 participated over six weekends to develop climate-policy recommendations. During the process, assembly members sought broader public feedback on draft recommendations through a project called ``The Climate Council asks Austria,'' which used Pol.is to collect and structure input from more than 5,000 Austrians.

From the Klimarat open data, we use three topic-specific Pol.is conversations: \emph{Production and Consumption}, \emph{Housing}, and \emph{Energy}. The Production and Consumption conversation asked participants for opinions and ideas on achieving climate neutrality by 2040, with discussion areas including product use and production, cost transparency, social and technological innovation, infrastructure and financing, and awareness-building and labeling. The Housing conversation asked for views on housing and climate neutrality, including the use of existing housing stock, renovation, adaptation needs, and emissions from construction and operation. The Energy conversation asked for views on reducing social energy consumption, shifting to renewable energy, increasing energy sufficiency, and financing the energy system, including carbon pricing. We use the statements from each topic as a separate subset in the DELTA analysis.

For each Pol.is conversation, we preprocess the data:
First, we retain only statements whose moderation field indicates inclusion. This is the platform's own moderation flag; it removes roughly half of each export, including off-topic or inappropriate statements and statements that were not voted on.
Next, we translate the statements (originally in Traditional Chinese for vTaiwan and German for the Klimarat conversations) into English using GPT-5, prompted to preserve hedging, sentiment, and policy terminology and to keep institutional names such as Klimarat as they are, so that remaining analyses are exclusively in English. Every statement DELTA scores is therefore a machine translation rather than the participant's original wording.
In the vTaiwan dataset, we additionally remove two kinds of statements by hand: factual self-disclosures, which report a fact about the speaker rather than making a proposal (``I have used Uber to hail a ride'', ``I have a driver's license for small cars''; 5 statements), and descriptive comparisons of Uber and taxis as they already operate (``I think UberX, on average, has better quality than taxis''; 22 statements). DELTA scores a statement by whether raters agree on which concrete actions it permits and rules out, so a purely descriptive claim gives a rater nothing to apply to a case. 
For the Klimarat conversations, we performed manual curation only to remove link spam and comments about the consultation process itself (two statements in Production and Consumption, one in Housing, none in Energy).

We apply the DELTA pipeline separately within each topic-specific subset, scoring every statement that survives cleaning. Case generation is seeded from 30 statements of each subset, favoring longer statements, and every statement is evaluated against 60 sampled cases for vTaiwan and 40 for each Klimarat conversation. The bridging analysis additionally restricts to statements with sufficient engagement from both opinion groups; \cref{appendix:bridging-sweep} details this threshold and shows that the results are not sensitive to it. The summary statistics are presented in \cref{tab:dataset-summary}.

\begin{table}[t]
    \centering
    \footnotesize
    \setlength{\tabcolsep}{3.5pt}
    \begin{tabular}{lrrrrrr}
        \toprule
        Dataset & $n$ & Cleaned & High-eng. & A & B \\
        \midrule
        Uber       & 199   & 72  & 63  & 410 & 398 \\
        Consump.   & 522   & 275 & 126 & 540 & 134 \\
        Housing     & 613   & 314 & 91  & 335 & 705 \\
        Energy     & 1,041 & 515 & 133 & 206 & 984 \\
        \bottomrule
    \end{tabular}
    \caption{Summary statistics by dataset. $n$ denotes the number of statements in the raw export. ``Cleaned'' denotes the number remaining after the moderation filter and manual curation; these are the statements DELTA scores. ``High-eng.'' denotes the number of high-engagement statements used in the bridging analysis, defined as those on which at least 20\% of the participants in each of groups A and B registered a response (agree, disagree, or pass). ``A'' and ``B'' denote the sizes of the two opinion groups algorithmically identified by Pol.is, estimated as the largest number of votes the group cast on any single statement.}
    \label{tab:dataset-summary}
\end{table}

\subsection{Engagement threshold}
\label{appendix:bridging-sweep}

\begin{figure}[h]
    \centering
    \includegraphics[width=\columnwidth]{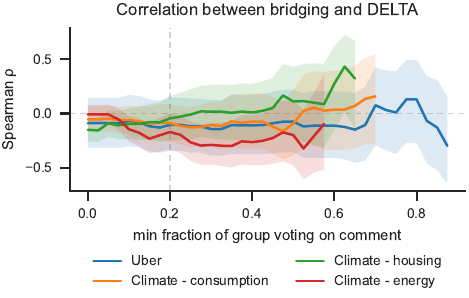}
    \caption{Spearman rank correlation between bridging, $\min(p_A, p_B)$,
    and DELTA as a function of the minimum engagement cutoff: statements are
    retained only if at least the given fraction of each opinion group's
    participants responded to them. Shaded bands are Fisher-$z$ $95\%$
    confidence intervals; lines truncate where fewer than $20$ statements
    remain. The dashed vertical line marks the $0.2$ cutoff used in
    \Cref{fig:bridging-delta}. The climate--energy correlation is negative
    at every cutoff and significantly so across a band of intermediate
    cutoffs; no dataset shows a stable positive correlation at any cutoff.}
    \label{fig:bridging-sweep}
\end{figure}

The group agreement rates $p_A$ and $p_B$ underlying the bridging analysis in
\Cref{fig:bridging-delta} are estimated from observed votes, Laplace-smoothed
as $p = (a+1)/(a+d+2)$, where $a$ and $d$ are the group's agree and disagree
counts on the statement; passes are not counted as votes. Some engagement
filter is necessary when using these estimates: Pol.is shows different
statements to different participants, so statements submitted late in a
conversation are seen by few voters, and rates estimated from a handful of
votes are noisy. We therefore retain only statements on which at least
$20\%$ of the participants in each opinion group registered a response
(agree, disagree, or pass), taking as each group's size the largest number of
votes it cast on any single statement (columns A and B of
\cref{tab:dataset-summary}). The value of $20\%$ itself, however, is a
judgment call.
\Cref{fig:bridging-sweep} therefore recomputes the Spearman rank correlation
between bridging and DELTA at every engagement cutoff from $0$ (no filter) to
$0.9$, with each line truncated once fewer than $20$ statements remain;
shaded bands are Fisher-$z$ $95\%$ confidence intervals, which widen as the
filter shrinks the sample.

The conclusions do not depend on the cutoff. Most importantly, at no cutoff
does any dataset show a stable positive correlation---the direction the
ambiguity-as-compromise hypothesis predicts. The vTaiwan--Uber and
consumption conversations hover around zero throughout ($\rho$ between
$-0.30$ and $+0.16$), with intervals that always include zero. The housing
conversation is weakly negative only before any engagement filtering is
applied ($\rho \approx -0.15$), consistent with noise from low-engagement
statements; its one positive interval, at a single isolated high cutoff, is
no more than chance would produce across the roughly $120$ intervals
plotted. The energy conversation is the exception: its correlation is
negative at every cutoff, and the interval excludes zero across the
contiguous band of cutoffs from $0.1$ to $0.4$ ($\rho$ between $-0.14$ and
$-0.30$), which includes the $0.2$ cutoff used in the main text. We treat
this as suggestive rather than conclusive: it appears in one dataset of
four, and the corresponding Pearson correlations are weaker
(\cref{fig:bridging-delta}), indicating an association carried by ordering
rather than by a linear trend in raw scores.

\section{Habermas dataset details} \label{appendix:Habermas}

The Virtual Citizens' Assembly in the Habermas study was a controlled deliberation experiment involving a demographically representative sample of the United Kingdom population~\cite{tessler2024ai}.
Participants were recruited through randomized stratified sampling (implemented by the Sortition Foundation) to ensure representativeness across age, gender identity, ethnicity, socioeconomic status, and geographic region. The assembly took place over three weekly one-hour sessions in April and May 2023. Our research use is consistent with their intention.

Unlike the convenience-sampled cohorts in the broader study, the Virtual Citizens' Assembly was specifically designed to simulate the deliberation phase of a real citizens' assembly, albeit without the typical fact-finding or expert testimony phases.
All participant groups deliberated over the same nine questions, enabling cross-group comparison of deliberative outcomes.
The nine assembly questions addressed potentially divisive issues in UK public policy.
They were selected from a larger corpus of approximately 5,500 deliberation questions based on participants' ratings of topic importance during earlier phases of data collection:
\begin{enumerate}[leftmargin=1em, labelindent=0pt, noitemsep]
	\item Should we be trying to reduce the number of people in prison?
	\item Should we raise the minimum wage in the UK?
	\item Should we raise the retirement age from 66 to 68?
	\item Should the UK change its asylum seeking policy to make it easier to enter the country?
	\item Should the UK apply to rejoin the European Union?
	\item Should we be doing more to encourage a greater degree of national pride?
	\item Should the government provide universal free childcare from birth?
	\item Has the Internet been a net positive for humanity?
	\item Should the UK set a goal to be carbon neutral (reach net zero) by 2050?
\end{enumerate}

One \emph{deliberation round} on a question refers to a group of 4-5 participants following a structured deliberation protocol:
\begin{enumerate}
	\item \textbf{Opinion Elicitation}: Participants were given five minutes to write a short paragraph (10--200 words) describing their personal opinion on the question. Groups consisted of five participants each.

	\item \textbf{Initial Statement Generation}: The Habermas Machine, an AI system based on large language models, generated four candidate ``group statements'' designed to capture common ground among the participants' opinions.

	\item \textbf{Rating and Ranking}: Participants independently rated each candidate statement on two dimensions: personal agreement (7-point scale from ``Strongly Disagree'' to ``Strongly Agree'') and statement quality (7-point scale from ``Terrible Quality'' to ``Excellent Quality''). They also provided preference rankings of all the statements.

	\item \textbf{Critique}: The top-ranked initial statement (selected via the Schulze method) was returned to participants, who wrote critiques identifying areas for improvement.

	\item \textbf{Revised Statement Generation}: The AI generated four revised candidate statements incorporating the critiques along with the original opinions.

	\item \textbf{Final Evaluation}: Participants rated, ranked, and compared the revised statements, then indicated their preference between the winning initial and revised statements.
\end{enumerate}
Each deliberation round lasted approximately 15 minutes, with participants completing three questions per session.

\subsection{Extraction and cleaning}

The original dataset from~\citet{tessler2024ai} is available \href{https://github.com/google-deepmind/habermas_machine/tree/main}{on Github}.
We use two of its tables. The candidate-comparisons table records, for every participant, round, and iteration, the candidate statements shown, the participant's agreement and quality rating of each candidate, and their preference ranking over the candidates. The position-statement-ratings table records each participant's agreement with a statement affirming the question, elicited before and after deliberation. In both tables a participant's judgments are stored as parallel arrays---candidate identifiers alongside numerical ranks and ratings---which we unnest jointly into one record per participant, round, iteration, and candidate. Categorical rating labels are mapped to the 1--7 scale; rankings a participant did not submit are stored by the release as $-1$, which we treat as missing, and rows from the release's \texttt{MOCK} control condition carry no agreement rating and are excluded from the agreement analysis.

We clean the extracted data as follows.
First, we exclude question 8 about the Internet: it asks for an evaluation rather than a prescription, so its statements have no policy content whose interpretation could diverge.
Second, we keep only deliberation rounds that completed both the initial and the revised iteration, leaving 309 rounds.
Third, we recompute each round's winning statement per iteration. The top-candidate field in the release is a placeholder, so we identify the winner as the candidate with the lowest mean numerical rank across the round's participants; candidates with no text or no rankings are ineligible, and ties fall to the earlier candidate, matching the order in which participants saw them. Six round-iterations contain no eligible candidate, and two rounds thereby lose both of their iterations, leaving 307 rounds and 612 winning statements.
Finally, we truncate 29 collective statements where it was clear that the raw text contained additional misidentified content,
for example, the statement
\begin{quotation}
Yes, we should be trying to reduce the number of people in prison. There is much evidence to suggest that prison can have a negative effect on prisoners, with no benefit to the wider community. Increasing emphasis should be placed on community-based punishment, rehabilitation and reintegration, with a reduction in the number of prison sentences given. There may be a need to build more prisons to reduce overcrowding, however we believe this should not be done at the expense of rehabilitation and reintegration. Prison should be used as a last resort for those who are a danger to society.

Final Comments from the group participants: The group worked really well together, sharing opinions and working to come to a consensus. We did not have anyone who strongly disagreed with the opinion of the majority of the group so it made the discussion and consensus easier.
How did this tool help participants to overcome their fear of working in groups? This tool helped participants to overcome their fear of working in groups as it helped them to get to know others in the group and their opinions, and to feel comfortable sharing their opinions.
\end{quotation}
was truncated to the first paragraph only.

The 612 winning statements---the highest-ranked draft and revised statements across all deliberation rounds---form the policy sets for the DELTA analysis: one set per question (72--82 statements each), scored in one run per question so that scores are only ever compared within a question. Each run used 30 raters and 60 sampled cases, with the generation and judgment settings of \cref{appendix:model-and-prompt}.
Alongside the winning statements we extract two participant-level tables: each participant's agreement rating of their round's winning statements (3,060 ratings, of which 2,350 carry an agreement value; the remainder are \texttt{MOCK} rows), and each participant's pre- and post-deliberation agreement with the question's affirming statement, which the figure below uses to identify initially divided rounds. The position ratings carry no round identifier in the release, so they are joined to rounds through the participant--question pair.
\Cref{tab:assembly_stats} presents summary statistics before and after cleaning.

\begin{table}[!h]
	\centering
	\begin{tabular}{lr}
		\toprule
		\textbf{Metric} & \textbf{Value} \\
		\midrule
		\multicolumn{2}{l}{\textit{Raw release (nine questions)}} \\
		\quad Deliberation questions & 9 \\
		\quad Unique deliberation rounds & 344 \\
		\quad Total participants & 1,720 \\
		\quad Total collective statements & 2,693 \\
		\quad\quad Draft statements & 1,367 \\
		\quad\quad Revised statements & 1,326 \\
		\midrule
		\multicolumn{2}{l}{\textit{After cleaning (eight questions)}} \\
		\quad Rds completing both iterations & 309 \\
		\quad Rds with a winning statement & 307 \\
		\quad Winning stmts (draft + revised) & 612 \\
		\quad Ratings of winning statements & 3,060 \\
		\quad\quad with an agreement value & 2,350 \\
		\midrule
		\multicolumn{2}{l}{\textit{Cleaned rounds / winning stmts per question}} \\
		\quad Prison population & 41 / 82 \\
		\quad Minimum wage & 41 / 81 \\
		\quad Asylum policy & 38 / 76 \\
		\quad National pride & 38 / 76 \\
		\quad Universal childcare & 38 / 75 \\
		\quad Retirement age & 38 / 76 \\
		\quad European Union & 37 / 74 \\
		\quad Carbon neutrality & 36 / 72 \\
		\bottomrule
	\end{tabular}
	\caption{Summary statistics for the raw release and the cleaned dataset. The raw block spans all nine questions; cleaning excludes the Internet question, rounds that did not complete both iterations, and round-iterations with no rateable winning statement. Where a question's statement count is below twice its round count, one iteration of one round produced no rateable winner.}
	\label{tab:assembly_stats}
\end{table}

\subsection{The agreement-distribution figure}

\Cref{fig:habermas-agreement} relates each participant's agreement with their round's winning statement to that statement's DELTA score.

First, as described in the main text, we restrict to rounds whose participants were initially divided: at least one member strictly above the neutral midpoint (positions 5--7) on the pre-deliberation position rating, and at least one strictly below (positions 1--3); participants exactly at the midpoint count toward neither side. A round that began in agreement cannot show ambiguity failing to bridge opposed positions. This restriction keeps 203 of the 307 rounds and 1,706 of the agreement ratings.

Second, DELTA scores are comparable only within a single scoring run, so the horizontal axis is the percentile rank of the statement's DELTA score within its question. This pools the eight questions onto a common $[0,1]$ axis without assuming that DELTA scores are similarly distributed across questions.

Third, for each level $k$ of the 1--7 agreement scale, we form the indicator that a rating equals $k$ and smooth its share as a function of DELTA rank with LOWESS (bandwidth fraction $0.45$, evaluated on a common grid). The $95\%$ confidence bands come from a cluster bootstrap with 300 resamples that resamples whole deliberation rounds with replacement, rather than individual ratings: all ratings in a round refer to the same winning statement and share its DELTA value, so their errors are correlated, and resampling individual ratings would understate the uncertainty.

\section{Model and prompts} \label{appendix:model-and-prompt}

\subsection{Case and rater generation}

To generate cases and raters, we use GPT-5 with medium thinking effort, temperature 1, and a maximum of 10{,}000 output tokens. The rater-expansion step, which appends one short sentence per call, instead runs with thinking disabled and a 200-token cap.

Cases are generated from \emph{pairs} of policies: the prompt in \cref{fig:prompt-generate-scenarios} presents two policies as alternative versions of the same rule and asks for three short fictional scenarios that comply with the first while violating the second. Because policies in a set typically express similar goals, this forces generation onto precisely the distinctions between them. For each dataset we sample 30 ordered pairs uniformly at random (fixed seed) from the dataset's seed policies, yielding a pool of 90 scenarios, from which the final case set is randomly subsampled: 60 cases for the vTaiwan and Habermas runs, 40 for the Klimarat runs.

Raters are generated in pairs with opposing interests using the top prompt in \cref{fig:prompt-generate-rater-description}, ten stakeholders (five pairs) per call; we run the query three times to obtain 30 rater identities per dataset. Generating in several smaller calls rather than one large one is deliberate, as a single call tends to produce a more homogeneous list. Each identity is then expanded using the bottom prompt in \cref{fig:prompt-generate-rater-description}, which takes the current description and adds one sentence to it; four applications yield the final five-sentence rater description.

To illustrate the outputs of this pipeline, we give a worked example drawn from the Habermas topic series (\cref{appendix:Habermas}), for the discussion question ``Should we be trying to reduce the number of people in prison?'' The policies in this run are the collective statements generated for that question, and the experiment configuration describes a case as a specific criminal justice policy or related action---for instance, a change to sentencing guidelines, rehabilitation programs, prison funding, or parole decisions---that raters can judge each statement against. One case generated from a pair of collective statements reads:
\begin{quote}
	Presumptive Non-Custodial Sentencing for Low-Level, Non-Violent Offences: Amend the Sentencing Code to create a statutory presumption against custodial sentences under 12 months, replacing them with structured community orders combining unpaid work, fines, curfews, and mandated treatment; judges may depart only with written reasons; cap custody for technical probation breaches at 7 days; increase probation staffing and electronic monitoring capacity; require quarterly compliance reporting to victims and courts; expected to reduce short-term prison receptions. Tradeoffs include higher probation caseloads and risk of non-compliance.
\end{quote}
An example rater identity generated for the same question is:
\begin{quote}
	This person is a suburban homeowner and taxpayer in Oakland County, Michigan, age 67 and on a fixed income, who supports decarceration only if a statutory budget lockbox dedicates documented savings to property tax relief, local victim services, and neighborhood patrols, with quarterly public dashboards on recidivism and costs and a five-year sunset requiring reauthorization; they care about fiscal accountability and visible community benefits.
\end{quote}
The four expansion passes then append one sentence each, grounding the identity in concrete experiences and increasingly specific positions on the question; for this rater they add:
\begin{quote}
	After her car was stolen from her driveway in Rochester Hills last summer, she’ll only support decarceration if verified savings show up as a line on her property tax bill and fund extra neighborhood patrols and victim services. After paying a \$500 insurance deductible and replacing her house locks when the thief took her spare key, she wants savings earmarked for victim reimbursements like deductibles and locksmith fees. Since her car was never recovered and no one was charged, she wants an audited county fund that reimburses deductibles and locksmith costs within 30 days, itemized as a credit on her property tax bill. She wants a hard trigger that pauses any decarceration plan if auto thefts or recidivism rise in her ZIP code for two consecutive quarters.
\end{quote}
Note that although the underlying deliberation took place in the UK, the generation prompts carry only the discussion question and the policy texts, so the model was free to situate stakeholders elsewhere.

\subsection{Judgment simulation}

\Cref{fig:prompt-judgment} contains the prompt used to simulate a rater's judgment of a policy against a case in the Pol.is and Habermas datasets. The university speech and adversarial datasets use the same structure with domain-specific wording (\cref{fig:prompt-judgment-validation}): raters judge whether a speech incident is permitted or prohibited under a university policy, and whether a proposed constitutional-reform implementation is consistent or inconsistent with a stated aspiration. Each judgment is elicited twice, under a positive and a negative framing of the final question, in separate queries; the two answers are collapsed into the revealed probability described in \cref{sec:ambiguity}.

We use Gemini-2.5-Flash for this task, with temperature 1 and thinking disabled, under a system prompt restricting the answer to yes or no. Each rater $\times$ policy $\times$ case $\times$ framing query is issued once, and the sampled answer is recorded as a binary value. A lightweight non-reasoning model is a deliberate choice beyond its low cost, which addresses the combinatorially large number of queries: the DELTA score is built on the spread of answers across raters, which reasoning tends to collapse.

\subsection{LLM-as-a-judge baselines}

The two direct LLM-as-a-judge baselines rate each policy on a 1--7 scale from its text alone, with no cases and no rater personas. The system prompt carries the rubric, and the user prompt supplies the discussion question and the statement to be rated. The first rubric (\cref{fig:prompt-ambiguity}) defines ambiguity in dictionary-style terms of clarity and specificity. The second rubric (\crefrange{fig:prompt-disagreement-1}{fig:prompt-disagreement-3}) mirrors the DELTA mechanism: it defines \emph{consequential ambiguity} as the disagreement that would arise among diverse stakeholders classifying concrete actions as consistent or inconsistent with the statement, provides guiding patterns and three worked examples, and instructs the model to mentally generate boundary-testing scenarios before rating.

Three judge models rate every statement under both rubrics: GPT-5 (medium thinking effort, temperature 1), GPT-5.2 (thinking disabled, temperature 0), and Gemini-3-Pro Preview (medium thinking effort, temperature 1). Each judge produces one rating per statement, and a statement's baseline score is the mean of the three judges' ratings. The baselines are applied to the campus speech policies (\cref{app:human-study}) and to the adversarial policies (\cref{app:adversarial-procedure}).

\begin{figure*}[htbp]
	\begin{tcolorbox}[colback=gray!5!white, colframe=gray!40!black,
		title=Prompt to generate cases]

		\textbf{System prompt:} \\
		Give your output as a JSON list of strings, where each string is one scenario.\\

		\textbf{User prompt:} \\
		Here are two versions of \{policy topic\}.\\
		Version 1: \{policy 1\}\\
		Version 2: \{policy 2\}\\
		Please come up with 3 short, fictional scenarios that comply with the first version but violate the second. Each scenario should be a \{case format description\}. They should be realistic and detailed enough to clearly illustrate the difference between the two statutes. They should also be creative and cover a diverse range of situations.\\
		\{case example\}\\
		Only output the scenarios, do not include any analysis, reasoning, or commentary.
	\end{tcolorbox}
	\caption{Prompt to generate cases, three at a time, from an ordered pair of policies. Placeholders in braces are filled from the dataset's experiment configuration.}
	\label{fig:prompt-generate-scenarios}
\end{figure*}

\begin{figure*}[htbp]
	\begin{tcolorbox}[colback=gray!5!white, colframe=gray!40!black, title=Prompt to generate diverse rater identities for a policy topic]
		\textbf{System prompt:} \\
		Give your output as a JSON list of strings, where each string is one stakeholder, beginning with the phrase `This person is a...'.\\

		\textbf{User prompt:} \\
		You are writing \{policy topic\} and interested in considering the perspectives of various stakeholders. Please make a list of 10 total stakeholders. Generate these stakeholders in pairs who have opposing interests in some aspect.\\
		Here is an example pair:\\
		\{example pair\}\\
		Be creative in generating the stakeholders and draw from diverse professional identities while keeping the descriptions succinct. Flatten the final output as a list of stakeholders without grouping them into pairs.
	\end{tcolorbox}
	\vspace{5pt}
	\begin{tcolorbox}[colback=gray!5!white, colframe=gray!40!black, title=Prompt to expand a rater description]
		\textbf{System prompt:} (empty)\\

		\textbf{User prompt:} \\
		Here is a brief description of a person who is a stakeholder regarding \{policy topic\}.\\
		Description: \{current rater description\}\\
		Please write a very short sentence expanding on the description in connection with the policy topic. Be realistic and consistent about their position on the issue.
	\end{tcolorbox}
	\caption{Top: prompt to generate rater identities, ten at a time in opposing pairs. Bottom: prompt to expand a rater description, which takes the current description and adds one sentence to it; it is applied four times, starting from the generated identity, to reach the final five-sentence description.}
	\label{fig:prompt-generate-rater-description}
\end{figure*}

\begin{figure*}[htbp]
	\begin{tcolorbox}[colback=gray!5!white, colframe=gray!40!black, title=Prompt to simulate rater judgments]
		\textbf{System prompt:} \\
		Answer ONLY `yes' or `no' only to the question below. Do not provide an explanation.\\

		\textbf{User prompt:} \\
		\{rater description\}\\

		They are reviewing opinion statements on the question: ``\{question\}''\\

		OPINION STATEMENT:\\
		\{policy\}\\

		They consider how the opinion statement would translate into real-world policy actions. A candidate proposal is given below:\\

		POLICY PROPOSAL:\\
		\{case\}\\

		After thinking carefully from their own point of view, they answer a yes/no question.\\
		Question: \{framing question\}
	\end{tcolorbox}
	\caption{The judgment prompt for the Pol.is and Habermas datasets. The final question takes one of two framings, asked in separate queries: the positive framing ``Does the intention of the policy proposal support/is compatible with the principles and beliefs described in the opinion statement?'' and the negative framing ``Does the intention of the policy proposal contradict/stand in opposition to the principles and beliefs described in the opinion statement?''}
	\label{fig:prompt-judgment}
\end{figure*}

\begin{figure*}[htbp]
	\begin{tcolorbox}[colback=gray!5!white, colframe=gray!40!black, title=Prompt to simulate rater judgments: university speech]
		\textbf{System prompt:} \\
		Answer ONLY `yes' or `no' only to the question below. Do not provide an explanation.\\

		\textbf{User prompt:} \\
		\{rater description\}\\

		They are evaluating how a university speech policy applies to a specific incident on campus.\\

		UNIVERSITY POLICY:\\
		\{policy\}\\

		INCIDENT:\\
		\{case\}\\

		After thinking carefully from their own point of view, they answer a yes/no question.\\
		Question: \{framing question\}
	\end{tcolorbox}
	\vspace{5pt}
	\begin{tcolorbox}[colback=gray!5!white, colframe=gray!40!black, title=Prompt to simulate rater judgments: adversarial]
		\textbf{System prompt:} \\
		Answer ONLY `yes' or `no' only to the question below. Do not provide an explanation.\\

		\textbf{User prompt:} \\
		\{rater description\}\\

		They are evaluating whether a proposed constitutional-reform implementation is consistent with a stated aspiration for Australia's head-of-state arrangement.\\

		CONSTITUTIONAL ASPIRATION:\\
		\{policy\}\\

		PROPOSED IMPLEMENTATION:\\
		\{case\}\\

		After thinking carefully from their own point of view, they answer a yes/no question.\\
		Question: \{framing question\}
	\end{tcolorbox}
	\caption{The judgment prompts for the university speech dataset (top) and the adversarial dataset (bottom). As in \cref{fig:prompt-judgment}, each judgment is asked under two framings in separate queries. University speech: ``Would this person say the speech in the incident is permitted under the policy?'' and ``Would this person say the speech in the incident is prohibited under the policy?'' Adversarial: ``Would this person say the proposed implementation is consistent with the constitutional aspiration?'' and ``Would this person say the proposed implementation is inconsistent with the constitutional aspiration?''}
	\label{fig:prompt-judgment-validation}
\end{figure*}

\begin{figure*}[htbp]
	\begin{tcolorbox}[colback=gray!5!white, colframe=gray!40!black, title=Prompt to rate the ambiguity of a policy]
		\textbf{System prompt:} \\
		You are an expert at evaluating the clarity and specificity of group opinion statements.\\

        Your task is to rate the AMBIGUITY level of a collective statement on a scale from 1 to 7:\\

        1 = Extremely clear and specific - The statement takes a definitive stance with concrete details\\
        2 = Very clear - The statement has a clear position with minor hedging\\
        3 = Mostly clear - The statement is understandable but has some vague elements\\
        4 = Moderately ambiguous - The statement balances multiple views, making the overall stance unclear\\
        5 = Quite ambiguous - The statement hedges significantly or avoids taking a clear position\\
        6 = Very ambiguous - The statement is vague and could be interpreted in multiple contradictory ways\\
        7 = Extremely ambiguous - The statement is so vague it conveys almost no clear position\\

        Respond with ONLY a single digit (1-7). Do not include any other text.\\

		\textbf{User prompt:} \\
		Rate the ambiguity level of this collective statement:\\

		Question being discussed: \{question\}\\

		Collective statement:\\
		``\{statement\}''\\

		Ambiguity rating (1-7):
	\end{tcolorbox}
	\caption{Prompt for the LLM-as-a-judge ambiguity baseline.}
	\label{fig:prompt-ambiguity}
\end{figure*}

\begin{figure*}[htbp]
	\begin{tcolorbox}[colback=gray!5!white, colframe=gray!40!black, title=Prompt to rate a policy's likelihood to create disagreement (part 1 of 3)]
		\textbf{System prompt:} \\
		Role: You are an expert analyst in policy implementation.\\
Task: You will be given a policy discussion question, and then a collective statement. Your task is to rate how **consequentially ambiguous** the statement is on a scale of 1 to 7.\\

What is consequential ambiguity?\\

**Consequential ambiguity** measures how much disagreement would arise among diverse stakeholders when trying to determine whether specific real-world actions are consistent with the statement.\\

This is distinct from linguistic ambiguity or vagueness. We are not asking whether the statement *sounds* clear or uses hedging language. We are asking what happens when someone tries to *apply* it --- when a regulator, a citizen, or an affected party must look at a concrete situation and decide: does this comply with what the statement says?\\

The core question is: If ten people with different values, professional roles, and life experiences were shown the same concrete implementation scenario, would they agree on whether it is consistent with this statement?\\

A statement has low consequential ambiguity when real-world scenarios can be classified as consistent or inconsistent with little disagreement, even among people who hold very different political views. A statement has high consequential ambiguity when realistic scenarios routinely produce split judgments --- not because people are confused, but because the statement's key terms or boundaries genuinely support multiple defensible readings that lead to opposite conclusions.\\

Guiding patterns\\

Here are some patterns that might help, but they are far from being exhaustive.\\

**Terms that sound specific but have contested scope.** Words like ``essential services,'' ``reasonable accommodation,'' ``normal wear and tear,'' ``adequate funding,'' or ``net-zero'' feel like they narrow the meaning, but different stakeholders operationalize them differently. A landlord and a tenant will draw the line of ``normal wear and tear'' in very different places, and both readings are defensible under the statement.\\

**Quantifiers or standards that lack a shared referent.** Phrases like ``significant investment,'' ``world-class,'' ``best available technology,'' or ``proportionate response'' gesture at a standard without anchoring it to anything measurable. Even when combined with specific-sounding mechanisms, they can leave the operative judgment wide open.\\
	\end{tcolorbox}
	\caption{Prompt for the LLM-as-a-judge disagreement baseline, which mirrors the DELTA mechanism (part 1 of 3).}
	\label{fig:prompt-disagreement-1}
\end{figure*}

\begin{figure*}[htbp]
	\begin{tcolorbox}[colback=gray!5!white, colframe=gray!40!black, title=Prompt to rate a policy's likelihood to create disagreement (part 2 of 3)]

**Temporal, causal, or scope boundaries that shift under pressure.** ``From the date of completion,'' ``during working hours,'' ``as a result of the policy,'' or ``within the scope of employment'' all sound like they mark a clean boundary. But real scenarios frequently straddle these boundaries, and reasonable people will disagree about which side a particular case falls on.\\

**Implicit tradeoffs between competing values.** Statements that invoke two desirable goals without specifying priority (``ensure safety while preserving individual liberty,'' ``promote innovation while protecting consumers'') create ambiguity at exactly the points where the goals conflict. Any concrete scenario that sits at the tension point will split raters along value lines.\\

**Terms whose meaning is stakeholder-dependent.** Some words have stable meanings within a professional community but diverge across communities. ``Clinical assessment'' means something specific to an emergency physician but something different to a hospital administrator tracking throughput metrics. When a statement uses such terms and the affected stakeholders span multiple communities, the same scenario will be classified differently depending on whose definition prevails.\\

Examples\\

**Example A** (Score: 1)\\
> ``The retirement age for state pension eligibility should be raised from 66 to 68 beginning in 2028 and applying to everyone with no exceptions.''\\

Rationale: Almost every concrete scenario --- When do I retire? Am I eligible? Does this apply to me? --- has a clear answer derivable from the statement. The terms are numeric and categorical. Stakeholders may strongly oppose the policy, but they would agree on what it means in practice. It would be difficult to construct a realistic scenario where reasonable people disagree about whether it is consistent with this statement.\\

**Example B** (Score: 6)\\
> ``All new commercial developments must achieve net-zero carbon emissions from the date of completion.''\\

Rationale: This reads like precise regulatory language with a clear standard. But the operative terms are contested in practice. If a developer purchases carbon offsets from a reforestation project in another country, does the building qualify as net-zero? What if the developer is operationally net-zero but not the tenants? If a mixed-use building has commercial units on the ground floor and residential above, does it count as a ``commercial development''? A green energy advocate and a property developer would look at the same project and reach opposite conclusions, both defensibly supported by the text.\\
	\end{tcolorbox}
	\caption{Prompt for the LLM-as-a-judge disagreement baseline, which mirrors the DELTA mechanism (part 2 of 3).}
	\label{fig:prompt-disagreement-2}
\end{figure*}

\begin{figure*}[htbp]
	\begin{tcolorbox}[colback=gray!5!white, colframe=gray!40!black, title=Prompt to rate a policy's likelihood to create disagreement (part 3 of 3)]

**Example C** (Score: 3)\\
> ``The government should invest more in renewable energy while ensuring energy bills remain affordable for low-income households.''\\

Rationale: The statement is directional --- it calls for more investment and affordability protections. Most concrete government actions can be classified fairly clearly: a new solar subsidy program is consistent, cutting renewable funding is not. The main source of ambiguity is ``affordable,'' where different stakeholders might draw the threshold differently. But the disagreement space is limited because the statement does not specify a particular mechanism or level of investment, which paradoxically makes it harder to find scenarios that are genuinely contested rather than simply underspecified. A progressive and a fiscal conservative might disagree about whether a specific bill goes *far enough*, but that is a disagreement about adequacy, not about consistency with the statement's direction.\\

How to reason about this\\

Before rating, mentally generate three or four concrete scenarios that test the boundaries of the statement. For each, ask: would a diverse panel agree on whether it is consistent with the statement? If you find yourself easily constructing cases where reasonable people split, the score should be higher. If edge cases are hard to construct or would still resolve clearly, the score should be lower.\\

Consider the size and motivation of dissenting groups. If you can identify specific interest groups who have strong reasons to interpret the statement differently from the common reading --- and whose interpretation is defensible rather than self-serving --- the score should be higher, in proportion to how widespread and legitimate that alternative reading is. A single contrarian reading does not make a statement highly ambiguous; systematic divergence across major stakeholder groups does.\\

Remember: you are not rating whether the statement is good policy, whether it is controversial, or whether you personally find it clear. You are rating how much disagreement would arise when diverse people try to classify concrete scenarios as consistent or inconsistent with the statement.\\

Respond with ONLY a single digit (1-7). Do not include any other text.\\

		\textbf{User prompt:} \\
		Rate the consequential ambiguity of this collective statement out of 7:\\

		Question being discussed: \{question\}\\

		Collective statement:\\
		``\{statement\}''\\

		Consequential ambiguity rating (1-7):
	\end{tcolorbox}
	\caption{Prompt for the LLM-as-a-judge disagreement baseline, which mirrors the DELTA mechanism (part 3 of 3).}
	\label{fig:prompt-disagreement-3}
\end{figure*}

\end{document}